\documentclass[journal]{IEEEtran}
\usepackage{cite}
\usepackage{graphicx}
\usepackage{subfigure}
\usepackage{amsmath}
\usepackage{amssymb}
\usepackage{amsthm}
\usepackage{xspace}
\usepackage[export]{adjustbox}
\usepackage{tabularx}
\usepackage[table,xcdraw]{xcolor}
\usepackage{multirow}
\usepackage{color}
\usepackage{algorithm}
\usepackage{hhline}
\usepackage{lineno}
\usepackage{colortbl}
\usepackage[]{algpseudocode}
\usepackage{array}
\usepackage{booktabs}
\usepackage{textcomp}
\usepackage{pifont}
\usepackage{url}
\usepackage{array}
\usepackage{pifont}

\newcolumntype{Y}{>{\centering\arraybackslash}X}

\newcommand{\cmark}{\ding{51}}

\makeatletter
\def\hlinewd#1{
	\noalign{\ifnum0=`}\fi\hrule \@height #1 \futurelet
	\reserved@a\@xhline}

\definecolor{srcolor}{rgb}{1,0,0}

\begin{document}
	
	\title{Memory-guided Image De-raining \\ Using Time-Lapse Data}
	
	\author{ Jaehoon Cho,~\IEEEmembership{Student Member,~IEEE,}
		Seungryong Kim,~\IEEEmembership{Member,~IEEE,}\\
		and Kwanghoon Sohn,~\IEEEmembership{Senior Member,~IEEE,}
		\thanks{Jaehoon Cho and Kwanghoon Sohn are with the School of Electrical and Electronic Engineering, Yonsei University, Seoul 120-749, South Korea (e-mail: \{{rehoon},khsohn\}@yonsei.ac.kr).}%
		\thanks{Seungryong Kim is with the Department of Computer Science and Engineering, Korea University, Seoul 02841, Korea. (E-mail: seungryong\_kim@korea.ac.kr).}%
	}

	\markboth{}%
	{Shell \MakeLowercase{\textit{et al.}}: Bare Demo of IEEEtran.cls for Journals}

	\maketitle
	\IEEEpeerreviewmaketitle
	
	\begin{abstract}
		This paper addresses the problem of single image de-raining, that is, the task of recovering clean and rain-free background scenes from a single image obscured by a rainy artifact. Although recent advances adopt real-world time-lapse data to overcome the need for paired rain-clean images, they are limited to fully exploit the time-lapse data. The main cause is that, in terms of network architectures, they could not capture long-term rain streak information in the time-lapse data during training owing to the lack of memory components. To address this problem, we propose a novel network architecture based on a memory network that explicitly helps to capture long-term rain streak information in the time-lapse data. Our network comprises the encoder-decoder networks and a memory network. The features extracted from the encoder are read and updated in the memory network that contains several memory items to store rain streak-aware feature representations. With the read/update operation, the memory network retrieves relevant memory items in terms of the queries, enabling the memory items to represent the various rain streaks included in the time-lapse data. To boost the discriminative power of memory features, we also present a novel background selective whitening (BSW) loss for capturing only rain streak information in the memory network by erasing the background information.
		Experimental results on standard benchmarks demonstrate the effectiveness and superiority of our approach.  
	\end{abstract}


	\section{Introduction}
	\IEEEPARstart{F}{or} images captured in rainy environments, the performance of numerous computer vision and image processing algorithms, such as object detection~\cite{kang2011automatic,fu2019lightweight,Li2019DerainBenchmark}, visual tracking~\cite{song2018vital}, or semantic segmentation~\cite{jiang2020multi,cho2020single}, is often significantly degraded. Image de-raining, aiming to restore a clean (or de-rained) image from the rain image, has thus attracted much attention from researchers in computer vision and image processing community as an essential pre-processing step.
	
	\begin{figure}
		\centering
		\renewcommand{\thesubfigure}{}
		\subfigure[(a) Input]{\includegraphics[width=0.325\linewidth]{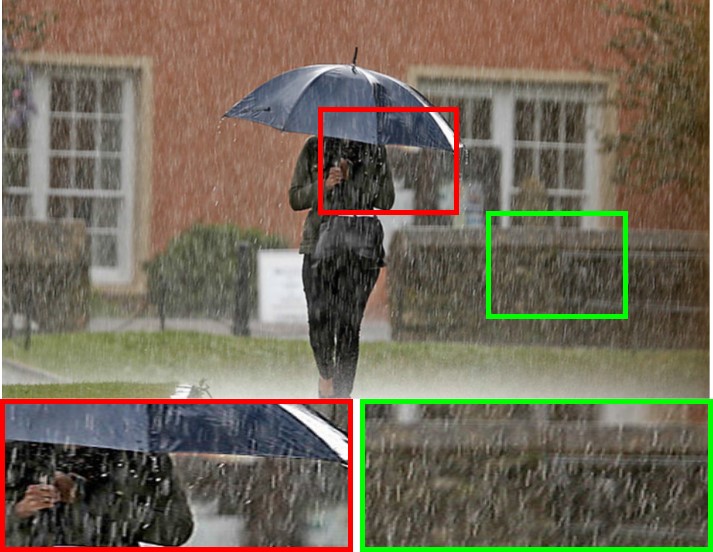}}
		\subfigure[(b) PReNet]{\includegraphics[width=0.325\linewidth]{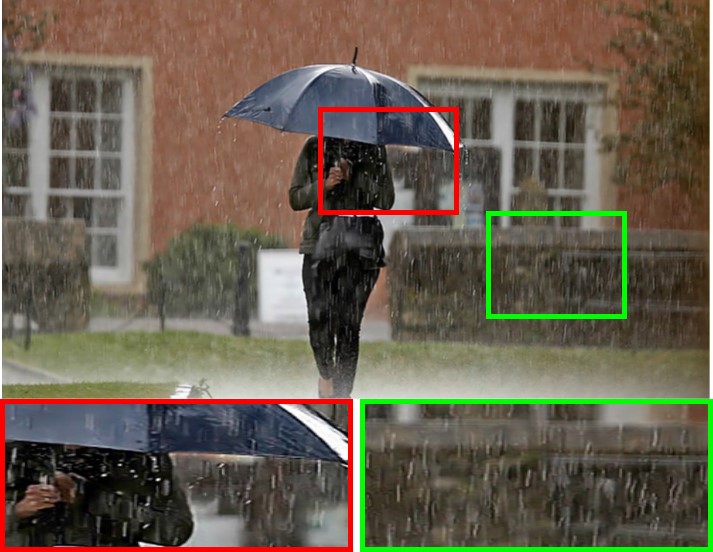}}
		\subfigure[(c) 
		MPRNet]{\includegraphics[width=0.325\linewidth]{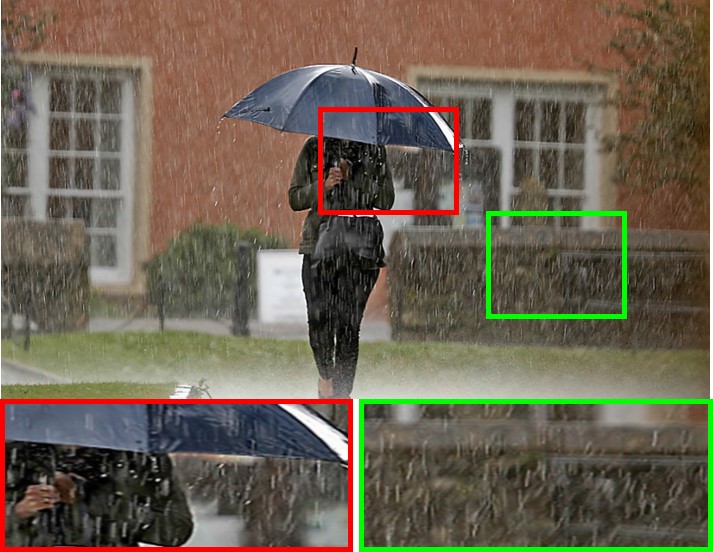}}
		\\ \vspace{-8pt}
		\subfigure[(d)
		SPANET]{\includegraphics[width=0.325\linewidth]{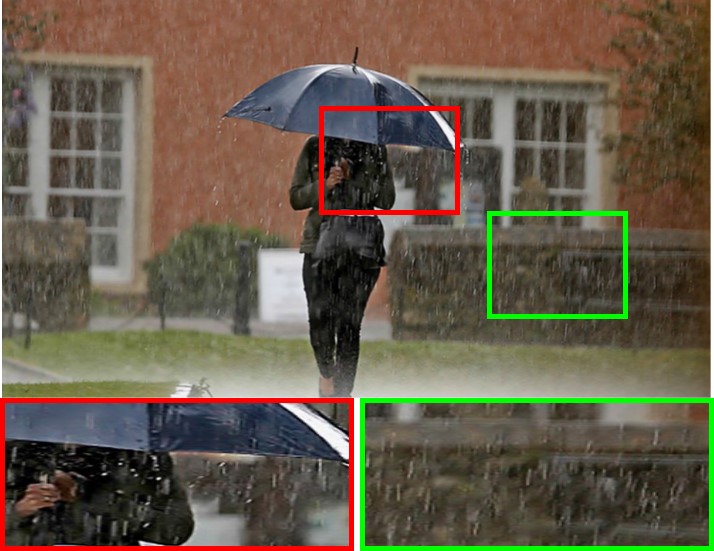}}
		\subfigure[(e)
		Cho \textit{et al.}]{\includegraphics[width=0.325\linewidth]{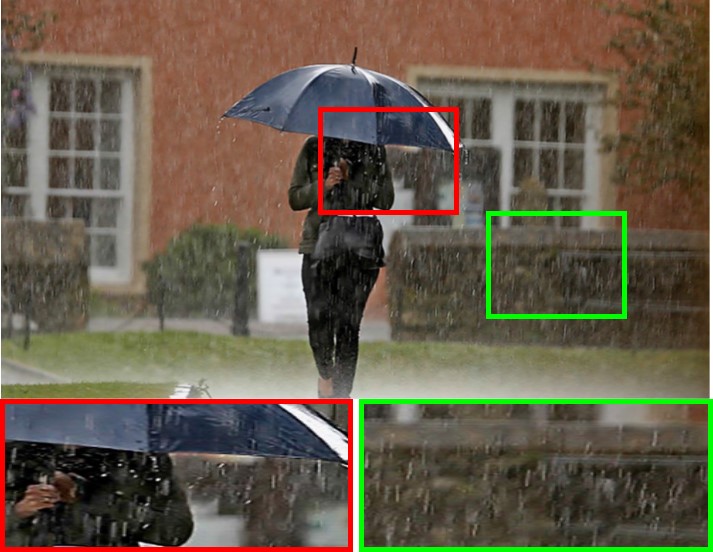}}
		\subfigure[(f)
		SIRR]{\includegraphics[width=0.325\linewidth]{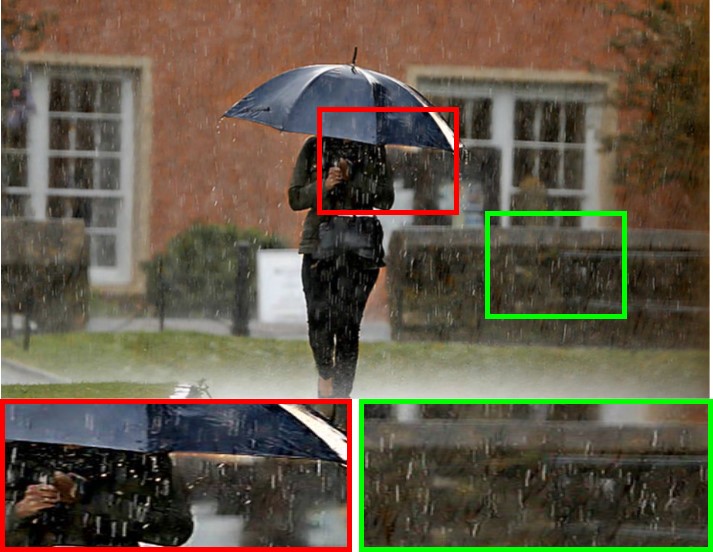}}
		\\ \vspace{-8pt}
		\subfigure[(g)
		Syn2Real]{\includegraphics[width=0.325\linewidth]{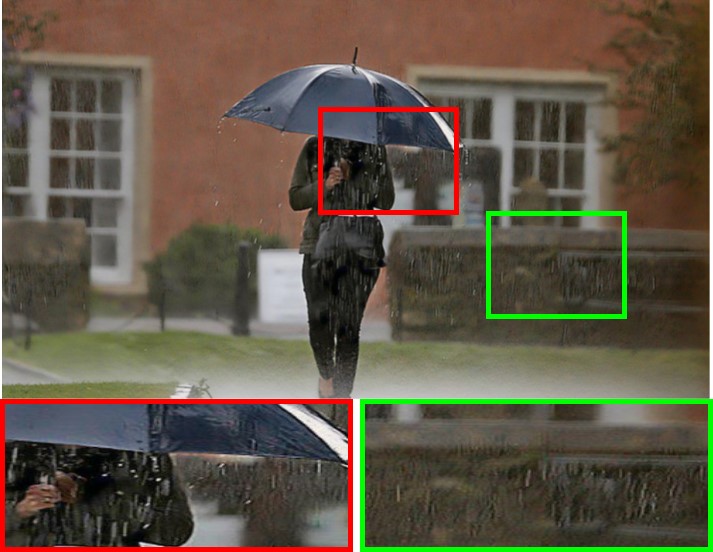}} 
		\subfigure[(h)
		MOSS]{\includegraphics[width=0.325\linewidth]{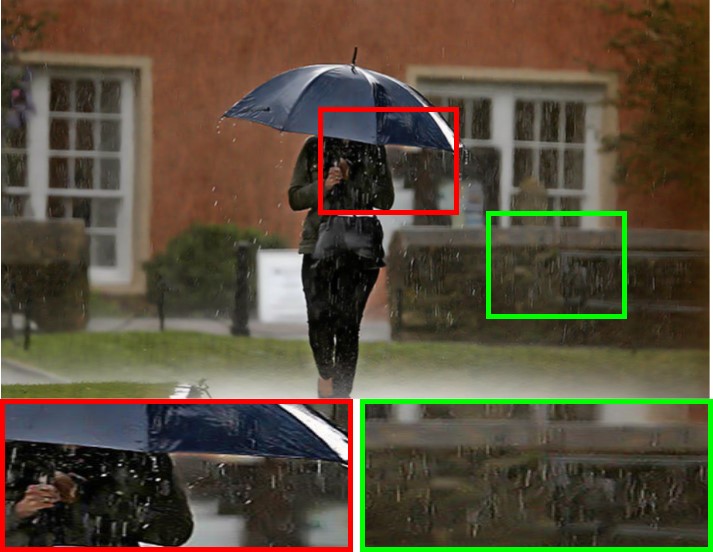}}
		\subfigure[(i)
		Ours]{\includegraphics[width=0.325\linewidth]{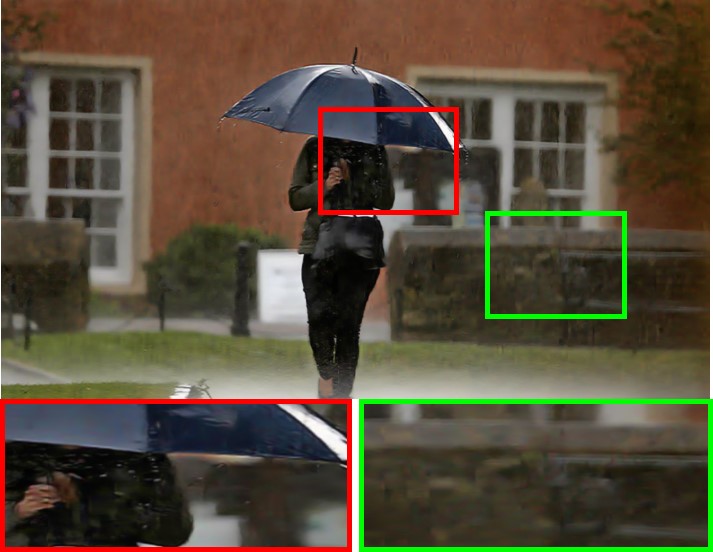}}
		\\ \vspace{-3pt}
		\caption{
			\textbf{Visual results of de-raining methods on real rain image degraded with various types of rain streaks.}
			(a) Input image, and results of (b) PReNet~\cite{ren2019progressive}, (c) MPRNet~\cite{zamir2021multi}, (d) SPANET~\cite{SPANet}, (e) Cho \textit{et al.}~\cite{cho2020single}, (f) SIRR~\cite{SIRR}, (g) Syn2Real~\cite{yasarla2020syn2real}, (h) MOSS~\cite{Huang2020memory}, and (i) Ours.
			Contrary to the existing methods,  the proposed method fully exploits the long-term rain streak information in the time-lapse data during training owing to the memory networks, thereby showing better performance.}
		\vspace{-10pt}
		\label{fig:1}
	\end{figure}

	With the significant success of deep convolutional neural networks (CNNs), many attempts been made to solve the image de-raining problem using deep CNNs~\cite{cho2020single,zhang2017image,li2018non,fu2017removing,fu2017clearing,zhang2018density,ren2019progressive,yang2020removing,yang2017deep,yang2019joint,SPANet,yang2019scale,wang2020model,zamir2021multi}. 
	Owing to the powerful feature representation of CNNs, these studies showed a significant performance improvement compared to conventional handcrafted methods~\cite{luo2015removing,li2016rain,chang2017transformed,zhu2017joint,gu2017joint}.
	Most existing methods using CNNs relied on synthetic rain images to train their networks because it is challenging to obtain paired real rain images and corresponding clean images. As easily expected, they exhibit limited performance when handling real rain images because the synthetic data cannot fully reflect various realistic rain streaks such as rain shape, direction, and intensity~\cite{yang2019joint,cho2020single,SPANet,yasarla2020syn2real}.
	
	To address this problem, several studies~\cite{SIRR,yasarla2020syn2real,Huang2020memory} have been developed to learn the rain streak priors with both synthesized paired data and unpaired real data.
	However, because they still rely on many synthetic rain images, they tend to fail when dealing with real rain images that have never been encountered during training~\cite{yang2020single}.
	The alternative direction is to utilize real-world \emph{time-lapse} data with constant background, except for time-varying rain streaks, because the consistent background information can be relatively easily modeled with the data. SPANET~\cite{SPANet} constructed a large-scale dataset of real rain/clean image pairs using time-lapse data. Cho \textit{et al.}~\cite{cho2020single} proposed a background consistency loss to estimate the consistent backgrounds of input images sampled from time-lapse data. They achieved improved performance for real-world rain images using real-world data. 
	Although they proved that using the time-lapse data for training enables overcoming the limitations of using synthetic data, their architectures are limited to fully utilizing the time-lapse data, because they rarely consider the long-term rain streak information across the time-lapse data.

	In this paper, we propose a novel network architecture and framework for single image de-raining based on a memory network that fully exploits the long-term rain streak information in time-lapse data. The key insight of our work is to store rain streak-relevant features in the memory network by removing consistent background information of time-lapse data.
	Specifically, our framework consists of the encoder--decoder networks and a memory network. The extracted features from the encoder, namely queries, are input into the memory network. Memory network, containing several memory items, stores and updates rain streak feature representations. This network explicitly retrieves relevant memory items with respect to queries, and thus memory items can represent various rain streaks included in the time-lapse data.  
	
	In addition, to capture only rain streaks-relevant features in the memory network, we propose a novel background selective whitening (BSW) to whiten the background information included in the queries so that the rain streak information is only stored in the memory network.
	These whitened queries allow the storage of diverse rain streaks such as direction, shape, density, and scale for different rain streaks into memory items without ground-truth paired data during training (update) and accessing them at test time (read).
	By incorporating our memory network with the BSW loss, the proposed method can effectively remove rain streaks and generate clearer background information.	
	
	Experimental results on several standard benchmarks, including synthetic datasets~\cite{fu2017removing,zhang2018density,yang2019joint} and real datasets~\cite{SPANet}, show the improvement of the proposed memory network and demonstrate the improved generalization ability on real data.


	Our main contributions are highlighted as follows.
	\begin{itemize}	
		\item 
		We propose a novel network architecture based on a memory network that stores and exploits long-term rain streak information in time-lapse data. 
		
		\item
		To encourage the memory network to capture only rain streak-relevant features, we introduce a novel BSW loss that removes the consistent background information in the time-lapse data.

		\item 
		We conducted extensive experiments on various datasets to demonstrate that the proposed approach outperforms recent state-of-the-art methods both quantitatively and qualitatively.
		

	\end{itemize}
	
	The remainder of this paper is organized as follows.
	Section II describes the related works.
	The proposed method is presented in Section III.
	Extensive performance validation is provided in Section IV, including an ablation study and comparison with the state of the art.
	Finally, Section V concludes this paper.

	\section{Related Work}

	\subsection{Single Image De-raining}
	
	In general, traditional single-image de-raining methods have been designed to explicitly model the physical characteristics of rain streaks.
	For instance, they adopted sparse coding~\cite{kang2011automatic}, dictionary learning~\cite{luo2015removing}, Gaussian mixture model~\cite{li2016rain}, and low-rank constraints~\cite{chang2017transformed}. Such methods often fail under complex rain conditions, and show over-smoothed images.
	
	By leveraging convolutional neural networks (CNNs), recent learning-based methods have been proposed in numerous studies.
	Fu \textit{et al.}~\cite{fu2017removing} first proposed de-raining network that decomposed the image into high- and low-frequency components and processed the high-frequency components using CNN.
	Yang \textit{et al.}~\cite{yang2017deep} estimated binary rain masks and rain, using a contextualized dilated network and they extended a prior work through a recurrent structure~\cite{yang2019joint}.
	Zhang and Patel~\cite{zhang2018density} presented a density-aware multi-stream de-raining network that uses a density cue from a rain density classifier.
	A recurrent framework leveraging a squeeze-and-excitation network~\cite{li2018recurrent}, a progressive network~\cite{ren2019progressive}, and a wavelet transform~\cite{yang2019scale} were introduced to gradually remove rain streaks.
	Wang \textit{et al.}~\cite{wang2020model} proposed a rain convolutional dictionary network in which the rain shapes were encoded.
	Yang \textit{et al.}~\cite{yang2020removing} designed fractal band learning network based on frequency band recovery.
	Hu \textit{et al.}~\cite{hu2019depth,hu2020single} proposed a depth-guided attention mechanism to remove rain and fog.
	Yasarla \textit{et al.}~\cite{yasarla2020confidence} proposed an image quality-based method that learns the quality or distortion level of each patch in the rain image. MPRNet~\cite{zamir2021multi} proposed a multi-stage architecture that progressively learns restoration functions for the degraded inputs.
	Several studies have used synthetic datasets and un-labeled real-world images to adapt real diverse rain streaks by designing an expectation maximization algorithm~\cite{SIRR} and the Gaussian process~\cite{yasarla2020syn2real}.
	More reviews of image de-raining methods are summarized well in~\cite{yang2020single}. 

	{Aforementioned approaches rely on large amounts of synthetic rain images for training, limiting their ability to handle real rain images.}
	Unlike the previous studies, we propose a novel architecture based on memory networks designed to fully exploit the long-term rain streak information in real-world time-lapse data.
	
	\subsection{Video De-raining}
	
	Early methods, such as those proposed by Garg and Nayar~\cite{garg2004detection,garg2005does,garg2007vision}, studied the visual effects of rain drops and developed a rain detection method using a physics-based motion--blur model.
	In addition, various video de-raining methods have been proposed to incorporate the spatial and temporal properties of rain streaks using \textit{k}-means clustering~\cite{zhang2006rain}, statistical frequencies~\cite{barnum2007spatio}, low-rank hypothesis~\cite{kim2015video,ren2017video}, tensor-based~\cite{jiang2017novel}, GMM~\cite{wei2017should}, and sparse coding~\cite{li2018video}.
	Recently, the CNN based methods have been investigated.
	Chen \textit{et al.}~\cite{chen2018robust} proposed a framework using superpixel segmentation to handle torrential rain with opaque rain occlusion.
	Liu \textit{et al.}~\cite{liu2018erase} designed a joint recurrent rain removal and reconstruction network that incorporates spatial texture appearances and temporal coherence. 
	Liu \textit{et al.}~\cite{liu2018d3r} developed	a DRR network to handle dynamic video contexts.
	Yang \textit{et al.}~\cite{yang2019frame} presented a two-stage recurrent network with dual-level flow regularization.
	Although they can leverage the temporal information by analyzing the difference between adjacent frames, these methods cannot be directly applied to single-image de-raining because of the lack of temporal knowledge.
	

	\subsection{Using Time-lapse Data}
	Various CNN-based methods adopt the time-lapse data to exploit the structure preservation properties in which background information is constant while temporal context changes over time, such as time-lapse video generation~\cite{nam2019end,anokhin2020high,cheng2020time} and intrinsic decomposition~\cite{ma2018single,lettry2018unsupervised,lettry2018deep}.
	Similar to our proposed method, several efforts have been dedicated to improving single-image de-raining~\cite{SPANet,cho2020single}. 
	SPANET~\cite{SPANet} constructed a large-scale real-world rain/clean paired dataset using time-lapse data and proposed a spatial attentive network (SPANet) that removed rain streaks in a local-to-global spatial attentive manner.
	Cho \textit{et al.}~\cite{cho2020single} introduced a large-scale time-lapse dataset and exploited the dataset for a estimating the consistent background without ground truth.
	They~\cite{SPANet,cho2020single} showed the benefits of exploiting the time-lapse data.
	However, their architectures were limited in their ability fully to exploit time-lapse data because they rarely consider the long-term rain streak information across time-lapse data.
	In contrast, our proposed method involves a novel network architecture based on memory networks, enabling the exploitation of the long-term rain streak information.
	
	\begin{figure*}
		\centering
		{\includegraphics[width=1\textwidth]{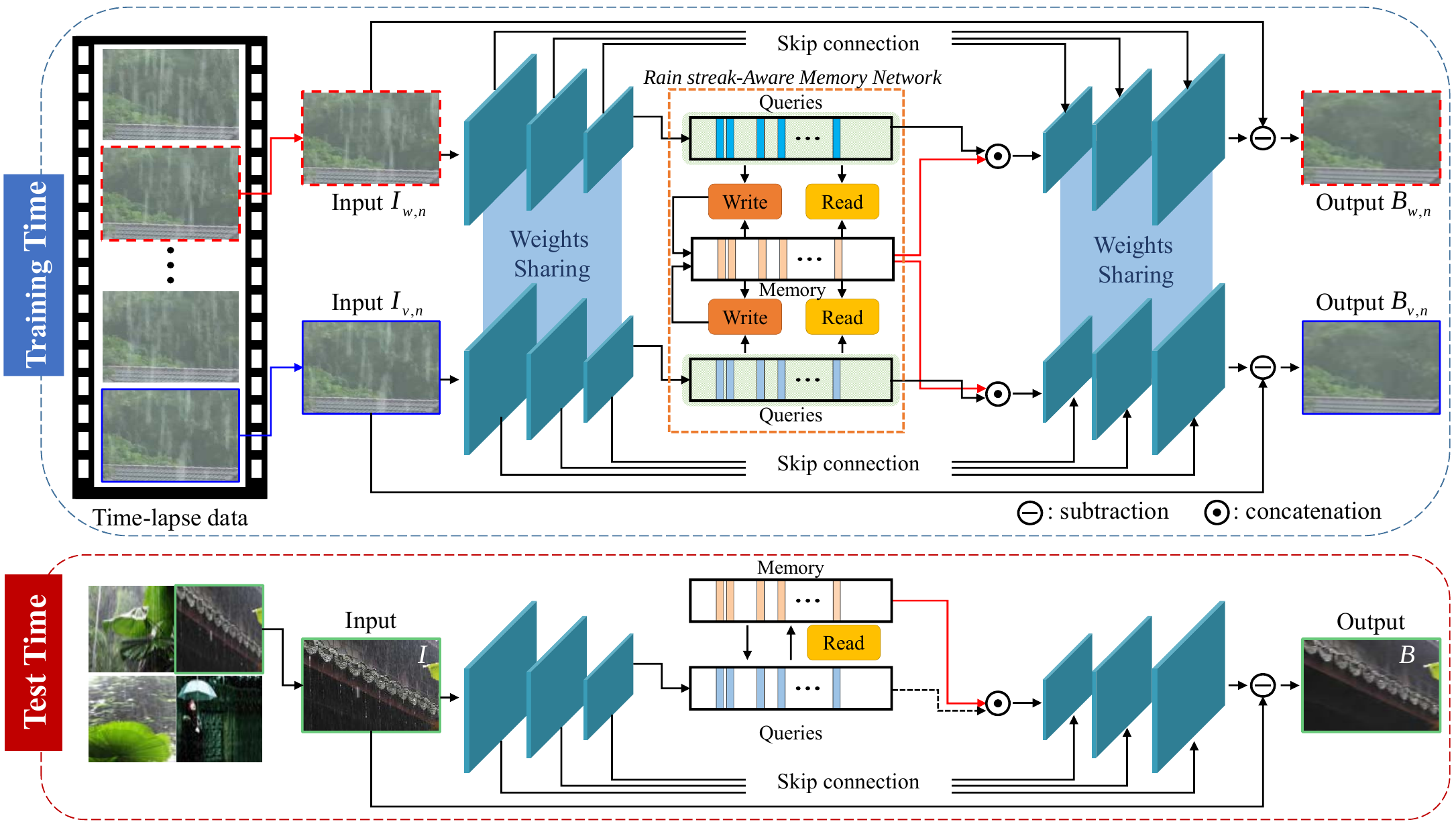}}
		\vspace{-8pt}
		\caption{
			\textbf{Overall network architecture.} 
			The encoder extracts the feature as a query $\mathbf{q}$. 
			The memory network performs read and update operations for reading and storing prototypical rain streak features and covers long-term information. 
			The retrieved memory items from the memory network and query are concatenated and then fed into a decoder to estimate the rain streaks. 
			The de-rained (clean) image $\hat{B}$ is obtained by subtracting the decoder output $R$ from the input rain image $I$.}
		\vspace{-10pt}
		\label{fig:2}
	\end{figure*}

	\subsection{Memory Networks}
	The memory network~\cite{weston2014memory,sukhbaatar2015end} is a trainable module that stores information in external memory and reads the relevant content from the memory.
	Weston \textit{et al.}~\cite{weston2014memory} first introduced a memory network.
	Graves \textit{et al.}~\cite{graves2014neural} introduced the application of external memory to extend the capability of neural networks.
	To record information more stably, Santoro \textit{et al.}~\cite{santoro2016one} proposed a memory-augmented neural network to rapidly update new data for a one-shot learning problem. 
	Owing to its flexibility, it has been widely adopted for solving various vision problems including movie understanding~\cite{na2017read}, video object segmentation~\cite{seong2020kernelized,oh2019video,lu2020video}, image generation~\cite{zhu2019dm}, VQA~\cite{fan2019heterogeneous}, and anomaly detection~\cite{gong2019memorizing,park2020learning}.
	
	Very recently, MOSS~\cite{Huang2020memory} was proposed as a de-raining method using memory networks.
	MOSS~\cite{Huang2020memory} is conceptually similar to the proposed approach in that it also adopts the memory network.
	However, our method differs from MOSS~\cite{Huang2020memory} for the following reasons.
	(i) 
	MOSS~\cite{Huang2020memory} often fails to estimate rain streaks because memory features are mixed with the background.
	In contrast, our method is more effective than previous works because of the newly proposed background whitening selective loss that removes the background information in the memory features.
	(ii) {In terms of the network architecture, while MOSS~\cite{Huang2020memory} that requires ground-truth data, our method is designed with a Siamese structure that can utilize the consistency between input images without ground-truth.}
	
	\section{Proposed Method}
	
	\subsection{Motivation and Overview}
	The single image de-raining task aims to remove the rainy effect and recover a rain-free background in a rainy image. An image degraded by a rainy artifact is generally formulated as
	\begin{equation}
		I = B + R,
	\end{equation}
	where $B$ and $R$ are the background or de-rained image and rain streak image, respectively. 
	Decomposing the image $I$ into $B$ and $R$ is notoriously challenging, because it is a highly ill-posed problem~\cite{yang2019joint,ren2019progressive,yang2020single}.
	
	To solve this problem, existing CNN-based methods~\cite{zhang2017image,li2018non,fu2017removing,fu2017clearing,zhang2018density,ren2019progressive,yang2019joint,SIRR,SPANet,yasarla2020syn2real,wang2020model} attempted to estimate a mapping function from $I$ to $B$ (or $R$). Most of the methods are formulated to only use synthetic rain/clean pairs as supervised signals, and they thus often fail to deal with real rain images that have never been encountered during training~\cite{yang2020single}.
	To address this issue, several studies~\cite{SIRR,yasarla2020syn2real,Huang2020memory,cho2020single} used both synthesized paired data and unpaired real data. Furthermore, SPANET~\cite{SPANet} and Cho \textit{et al.}~\cite{cho2020single} exploited real-world time-lapse data for training. 
	Although they apparently showed improved generalization ability on real data, their network architecture was limited to fully exploiting rain streak information shown in the time-lapse data during training because of the lack of long-term memory components.
	
	To overcome this issue, we propose a novel network architecture based on a \emph{memory} network that, during training, fully makes use of long-term rain streak information in the time-lapse data, as shown in Fig.~\ref{fig:2}.
	Specifically, our framework consists of the encoder--decoder networks and memory network. Our memory network contains several memory items, each of which stores rain steak-aware feature representations. The features from the encoders, namely queries, are used to read and update the rain steak-aware features in the memory. The decoders then take them as inputs to estimate the final rain streaks. This memory network helps to \emph{explicitly} consider the long-term rain streak information across the time-lapse data, thereby improving time-lapse data utilization for training.
	For the memory network to only contain rain-streak-relevant features while erasing the background information in queries, which improves the discriminative power of memory item features, we further present a novel background selective whitening (BSW) loss function, inspired by~\cite{choi2021robustnet}.
	The proposed loss disentangles the covariance extracted from the queries into the encoded background and rain streaks information, and then selectively suppresses only the background covariance.
	
	Specifically, as training data, we use a set of time-lapse data $\mathbf{D} = \{D_{n}\}_{n=1, ...,N}$ and ${D}_{n}=\{{I}_{t,n}\}_{t=1, ...,T}$, where $N$ is the number of all time-lapse data, $T$ is the total time of a time-lapse data, $n$ denotes the index of scenes, and $t$ denotes the time.
	We denote by $I_{t,n}$ and $\mathbf{q}_{t,n}$ an input rain image and a corresponding feature, respectively, at time $t$ in the $n^{th}$ scene. 
	Our objective is to infer a de-rained image $B_{t,n}$ for each $I_{t,n}$ through the proposed method.
	

	\begin{table}[ht]
		\centering
		\caption{
			\textbf{The details of encoder--decoder network:} C\_in and C\_out denote the number of channels of the input and output features, respectively. 
			Note that $\hat{\mathbf{p}}$ is the retrieved memory items.
			$\{ { \cdot {\rm{,}} \cdot } \}$ denotes the concatenation operator.}
		\vspace{-5pt}
		\renewcommand{\tabcolsep}{5mm}
		\resizebox{!}{0.18\textheight}{
			\begin{tabular}{cccc}
				\toprule
				\multicolumn{4}{c}{Encoder}\tabularnewline
				\midrule
				Layer(Output)       & C\_in & C\_out  & Input \\
				\midrule\midrule
				conv\_E1a     & 3     &  64      & $I$ \\
				conv\_E1b     & 64    &  64         & conv\_E1a \\
				pool\_E1      & 64    &  64     & conv\_E1b \\
				\midrule
				conv\_E2a     & 64    &  64      & pool\_E1 \\
				conv\_E2b     & 64   &  64     & conv\_E2a \\
				pool\_E2      & 64   &  64     & conv\_E2b \\
				\midrule
				conv\_E3a     & 64   &  64   & pool\_E2 \\
				conv\_E3b     & 64   &  64    & conv\_E3a \\
				pool\_E3      & 64   &  64   & conv\_E3b \\
				\midrule
				conv\_E4a     & 64   &  64      & pool\_E3 \\
				conv\_E4b     & 64   &  64   & conv\_E4a \\
				\midrule
				\multicolumn{4}{c}{Decoder}\tabularnewline
				\midrule
				Layer(Output)       & C\_in & C\_out   & Input \\
				\midrule\midrule
				conv\_D4a     & 128    &  64   & \{conv\_E4b, $\hat{\mathbf{p}}$\} \\
				conv\_D4b     & 64     &  64  & conv\_D4a \\
				upconv\_D3     & 64     &  64   & conv\_D4b \\
				\midrule
				conv\_D3a     & 128     &  64   & \{upconv\_D3, conv\_E3b\} \\
				conv\_D3b     & 64     &  64  & Conv\_D3a \\
				upconv\_D2     & 64     &  64  & Conv\_D3b \\
				\midrule
				conv\_D2a     & 128     &  64 & \{upconv\_D2, conv\_E2b\} \\
				conv\_D2b     & 64     &  64    & conv\_D2a \\
				upconv\_D1     & 64     &  64  & conv\_D2b \\
				\midrule
				conv\_D1a     & 128     &  64   & \{upconv\_D1, conv\_E1b\} \\
				conv\_D1b     & 64      &  64   & conv\_D1a \\
				Output     & 64      &  3     & conv\_D1b \\
				\bottomrule
			\end{tabular}}
			~\label{tab:NetArch}
			\vspace{-13pt}
		\end{table}

		\subsection{Network Architecture}
		In this section, we begin with a description of the proposed network architecture. We largely follow the encoder--decoder network, which has been widely adopted for existing single image de-raining~\cite{li2018non,zhang2018density,cho2020single,yasarla2020confidence}.	
		We describe a detailed description of our network architecture in Table~\ref{tab:NetArch}. 
		Specifically, for all convolutional layers, the kernel size was set to $3\times3$. In the encoder, all max-pooling layers have a kernel size and stride set to $2\times2$, resulting in the output features being down--scale by a factor of 2.
		The encoder inputs a rain image $I_{t,n}$ and then extracts the feature $\mathbf{q}_{t,n}\in \mathbb{R}^{H \times W \times C}$, which can be used as a query for the memory network, where $H$, $W$, and $C$ are the height, width, and number of channels, respectively. We denote by $\mathbf{q}^k_{t,n}\in\mathbb{R}^{C}$ for $k=1, ..., K$, where $K=H \times W$, each query of size $1 \times 1 \times C$ in the query map. The query is then inputted to the memory network to read or update the memory items, such that it records prototypical rain streak information. Note that, for simplicity, the subscripts $t$ and $n$ are omitted because the proposed network repeats the same process for each input image in the memory network. Detailed descriptions of the memory network are presented in the following sections.
		In the decoder, each layer is composed of $3 \times 3$ deconvolution and convolution layers followed by ReLU, which is connected to the encoder using skip connections.
		The deconvolution layer, implemented with transposed convolutional layers, has an upscaling factor of 2.
		The decoder inputs the retrieved memory items from the memory network and encoded features $\mathbf{q}$ to produce rain $\hat{R}$. Finally, the de-rained image $\hat{B}$ is obtained by subtracting $\hat{R}$ from input rain image $I$. Next, we describe how the memory network captures the rain streak information through the read and update process.

		\begin{figure*}
			\centering
			\renewcommand{\thesubfigure}{}	
			\subfigure[Real Rain Image]{\includegraphics[width=0.19\linewidth,height=0.07\textheight]{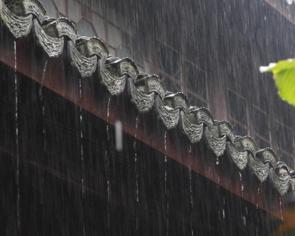}}
			\subfigure[(a)]{\includegraphics[width=0.19\linewidth,height=0.07\textheight]{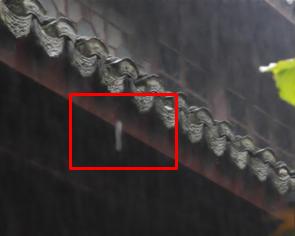}}
			\subfigure[(b)]{\includegraphics[width=0.19\linewidth,height=0.07\textheight]{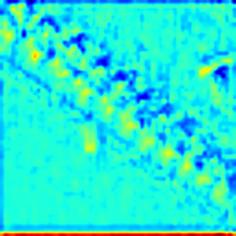}}
			\subfigure[(c)]{\includegraphics[width=0.19\linewidth,height=0.07\textheight]{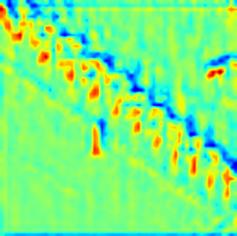}}
			\subfigure[(d)]{\includegraphics[width=0.19\linewidth,height=0.07\textheight]{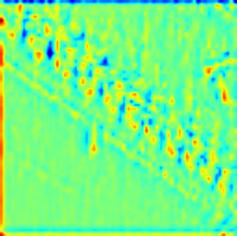}}\\
			\vspace{-8pt}
			\subfigure[]{\includegraphics[width=0.19\linewidth,height=0.07\textheight]{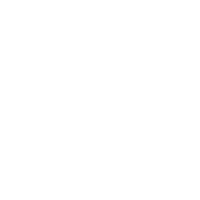}}
			\subfigure[(e)]{\includegraphics[width=0.19\linewidth,height=0.07\textheight]{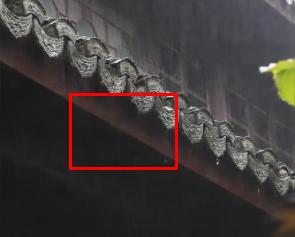}}
			\subfigure[(f)]{\includegraphics[width=0.19\linewidth,height=0.07\textheight]{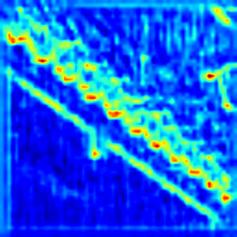}}
			\subfigure[(g)]{\includegraphics[width=0.19\linewidth,height=0.07\textheight]{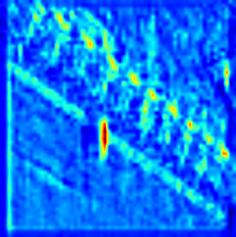}}
			\subfigure[(h)]{\includegraphics[width=0.19\linewidth,height=0.07\textheight]{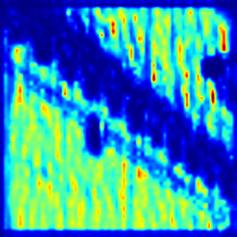}}\\
			\vspace{-8pt}
			\caption{
				\textbf{Visualization of de-rained results and memory feature maps trained with/without BSW loss.}
				(a) De-rained result trained without the BSW loss, (b)--(d) memory feature maps trained without BSW loss, (e) de-rained result trained with the BSW loss, and (f)--(h) memory feature maps trained with BSW loss.
				The proposed loss selectively removes the background information so that only rain streaks are stored in the memory network. 
				Note that the memory feature maps are up--sampled to match the image size (best viewed in color).
			}
			\label{fig:5}
			\vspace{-8pt}
		\end{figure*}

		\subsection{{Rain Streak-Aware Memory Network}}
		In our encoder-decoder networks, we design a memory network containing $M$ memory items to record various prototypical rain streak information and contain long-term rain streak information in time-lapse data. The memory network allows the storage and reading of diverse shapes such as scales, densities, and directions for different rain streaks into memory items during training, while accessing them at inference time.
		We denote by $\mathbf{p}_{m} \in \mathbb{R}^{C}$ for $m=1, ..., M$ the item in a memory network.
		
		\subsubsection{Read}
		To read the appropriate rain streak information in an input rain image, we first compute the similarity between query $\mathbf{q}^{k}$ and all memory items $\mathbf{p}_{m}$, resulting in a read weight matrix $\alpha^{k,m} \in \mathbb{R}^{M \times K}$ as follows:
		\begin{equation}\label{eq:2} 
			\alpha_{}^{k,m} = \frac{\exp (d(\mathbf{p}_{m},\mathbf{q}_{}^{k}))}{\sum\nolimits_{m' = 1}^M {\exp (d(\mathbf{p}_{m'},\mathbf{q}_{}^{k})) }},
		\end{equation}	
		where $d(\cdot,\cdot)$ is defined as a cosine similarity:
		\begin{equation}\label{eq:3}
			d(\mathbf{p}_{m},\mathbf{q}_{}^{k}) = \frac{\mathbf{p}_{m}^{\mathsf{T}}\mathbf{q}_{}^{k}}{{\left\| \mathbf{p}_{m} \right\|\left\| \mathbf{q}_{}^{k} \right\|}},
		\end{equation}
		where $\mathsf{T}$ represents the transpose operator.
		For each query $\mathbf{q}_{}^{k}$, we read the memory items by taking a weighted average of the memory items $\mathbf{p}_{m}$ with the corresponding weights $\alpha_{}^{k,m}$, and obtain the retrieved features ${\bf{\mathbf{\hat{p}}}}_{}^k\in \mathbb{R}^{C}$ as follows:
		\begin{equation}\label{eq:4}
			{{\bf{\mathbf{\hat{p}}}}_{}^k} = \sum\nolimits_{m'=1}^M {{\alpha_{}^{k,m'}}} {{\bf{p}}_{m'}}.
		\end{equation}	
		This step is repeated for input rain images sampled by time-lapse data. 
		Through this step, we can fully utilize diverse rain streaks containing all items. 
		
		By applying the reading operator to each query, we obtain a transformed feature map ${\bf{\mathbf{\hat{p}}}}_{}^k \in \mathbb{R}^{H \times W \times C}$. We concatenate the transformed feature map with the query along the channel dimension and input it to the decoder. This enables the decoder to estimate rain $\hat{R}$ using various rain streak information in the memory items. With the memory network, as shown in Fig.~\ref{fig:5} (b)--(d), various types of rain streaks can be captured by each memory items\footnote{Note that feature maps are up--sampled to match the image size.}.
		
		\subsubsection{Update}	
		To enhance the memory items during training, we dynamically select and store rain streak-relevant features into the memory network. Similar to the read operation, we compute an updated weight matrix $\beta_{}^{k,m} \in \mathbb{R}^{M \times K}$ between $\mathbf{p}_{m}$ and $\mathbf{q}_{}^{k}$:
		\begin{equation}\label{eq:5}
			\beta_{}^{k,m} = \frac{\exp (d(\mathbf{p}_{m},\mathbf{q}_{}^{k}))}{\sum\nolimits_{k' = 1}^K {\exp (d(\mathbf{p}_{m},\mathbf{q}_{}^{k'})) }},
		\end{equation}
		where we apply the softmax function along the $\mathbf{q}$-direction, as opposed to (\ref{eq:2}).
		The updated weight matrix $\beta$ is used to assign the extracted rain streak-relevant features $\mathbf{q}_{}$ to the {relevant} memory item. The memory items $\mathbf{p}_{m}$ are updated using $\mathbf{q}_{}$ weighted by $\beta$ as follows:
		\begin{equation}\label{eq:6}
			{\mathbf{\hat{p}}_{m}} = \mathrm{Norm}_\mathrm{L2}({{{\bf{p}}_m} + \sum\nolimits_{k'=1}^{K} {\beta_{}^{k,m}\mathbf{q}_{}^{k'}} }),
		\end{equation}
		where $\mathrm{Norm}_\mathrm{L2}$ denotes the $L_{2}$ norm.
		We utilize $\mathbf{q}_{}^{k'}$ to update $\mathbf{p}_{m}$. We train the memory items with a large number of real-world time-lapse data, enabling the most representative and discriminative rain streak-relevant features to be stored.
		
		During training, we access only a set of real-world time-lapse data without any ground truth to update the memory items assigned to store diverse rain streaks. At inference time, we compute $\mathbf{p}_{m}$ for all memory items without considering prior information such as rain density levels~\cite{zhang2018density} and a binary map indicating rain streak regions~\cite{yang2019joint,yang2020single}, and retrieve the rain streaks using Eq.~\ref{eq:2} and Eq.~\ref{eq:4}. Because our memory network is trained using diverse real-world time-lapse data, this strategy works well on rain images.

		\subsection{Loss Functions}
		Following Cho \textit{et al.}~\cite{cho2020single}, we adopt several loss functions to learn the our network architecture, including the memory network. In addition, the proposed method introduces a novel background selective whitening loss to remove the background information between the each queries to improve the discriminative power of the memory items.
		
		\subsubsection{Background Prediction Loss}
		Background prediction loss encourages the generation of consistent background images across time-lapse data.
		It is formulated as the $L_{1}$ distance between the estimated background images from time-lapse data, but at different times such that
		\begin{equation}\label{eq:}
			{{\mathcal{L}}_{b}}={\sum\limits_{n\in N} {\sum\limits_{\{w,v\}\in T}}} {\sum\limits_{i} {{{{\left\| {\hat{B}_{w,n}(i)}-{\hat{B}_{v,n}(i)} \right\|}_{1}}}}},	
		\end{equation} 
		where $\hat{B}_{w,n}$ is a background image that is decomposed from $I_{w,n}$.
		${w,v}$ represents the different times in $T$, and $n$ denotes the indexes of the scene.
		Here, ${\hat{B}_{w,n}(i)}$ and ${\hat{B}_{v,n}(i)}$ are the values at pixel $i$ from the image ${\hat{B}_{w,n}}$ and ${\hat{B}_{v,n}}$, respectively.
		
		\subsubsection{Cross Information Loss}
		This loss function is designed to encourage the estimated backgrounds to be close to the input images based on the assumption that the overall structure of the estimated backgrounds should be approximated well by input images~\cite{lettry2018unsupervised,lettry2018deep}.
		This loss helps to understand the structure of the overall layout information.
		It measures the $L_{1}$ distance between the estimated background image and the input image such that
		\begin{equation}\label{eq:cross}
			{{\mathcal{L}}_{c}}={\sum\limits_{n\in N} {\sum\limits_{\{w,v\}\in T}}} {\sum\limits_{i}{{{{\left\| {I_{w,n}(i)}-{\hat{B}_{v,n}(i)} \right\|}_{1}}}}}.
		\end{equation}
		Moreover, this loss function allows the network to produce good initial results during the early training phase.
		
		\subsubsection{Self Consistency Loss}
		This loss makes the summation of the estimated $\hat{B}$ and $\hat{R}$ to be the input image ${I}$ such that
		\begin{equation}\label{eq:self}
			{{\cal L}_s} = {\sum\limits_{n\in N} {\sum\limits_{w\in T}}} {\sum\limits_{i} {{\left\| {{I_{w,n}(i)} - ({\hat{B}_{w,n}(i)} + {\hat{R}_{w,n}(i)}) }\right\|}{_1}}},
		\end{equation}
		which acts as a regularizer.

		\begin{figure}
			\renewcommand{\thesubfigure}{}
			{\includegraphics[width=1\linewidth]{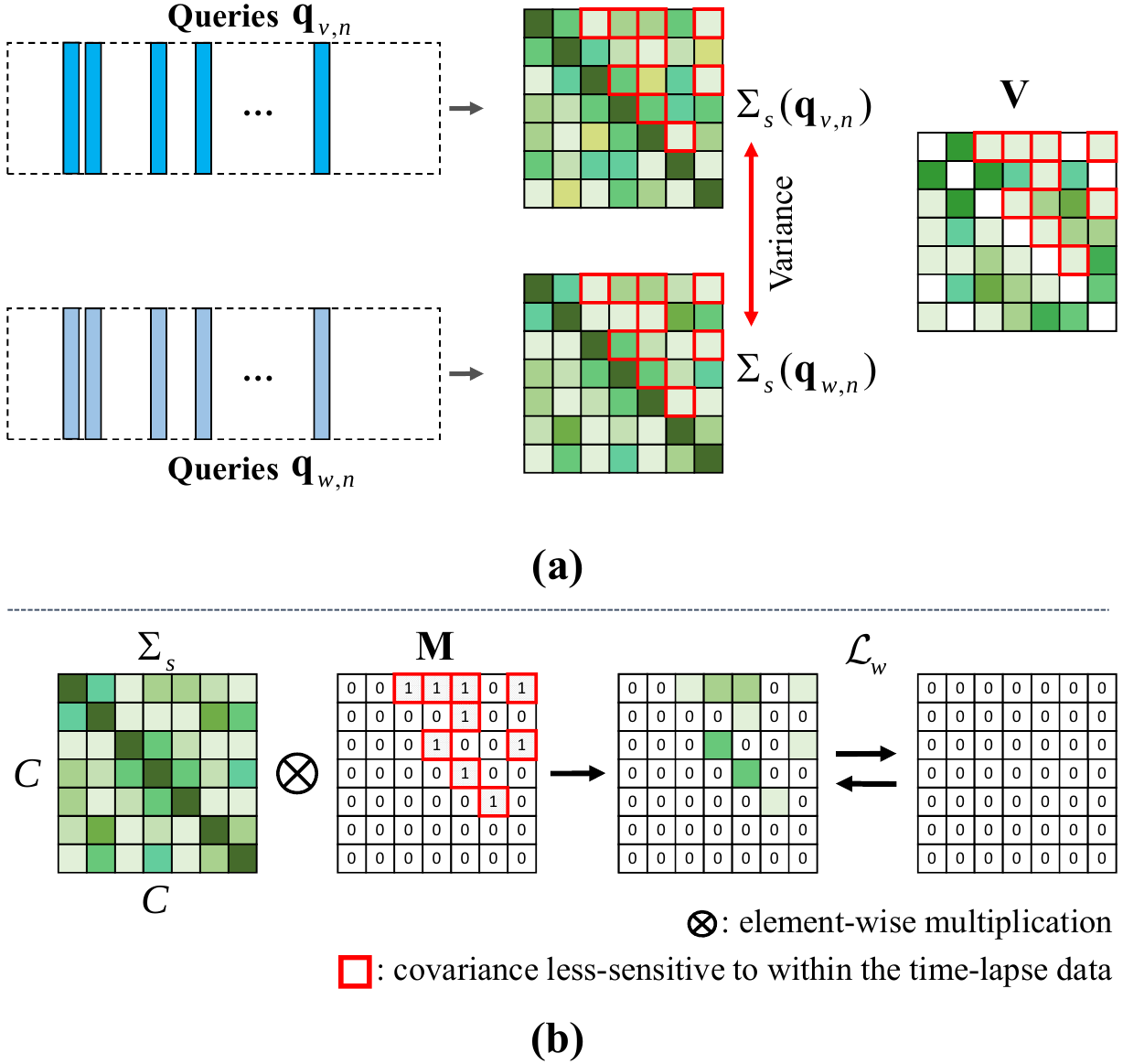}}
			\caption{
				\textbf{Illustration of the background selective whitening (BSW) loss.}
				(a) The variance matrix $\mathbf{V}$ is computed from the covariance matrices of each queries ($\mathbf{q}_{v,n},\mathbf{q}_{w,n}$) to identify the same elements (red boxes).
				Note that these matrices are symmetric. 
				(b) The covariance matrix $\Sigma_s$ is masked by the matrix $\mathbf{M}$, which belongs to a low variance value, to selectively suppress the same background covariances by the BSW loss $\mathcal{L}_{w}$.
			}
			\label{fig:3}
			\vspace{-8pt}
		\end{figure}

		\subsubsection{Background Selective Whitening Loss}
		In this section, we introduce a novel additional loss function, called background selective whitening (BSW) loss, to utilize only rain streak information in the memory networks, by erasing the background information in time-lapse data.
		We use the covariance extracted from the queries to separate them into the encoded background and rain streak. We then propose to handle only the background information, which can be selectively removed, thus improving the de-raining performance.
		As shown in Fig.~\ref{fig:3}, each queries is separated into two groups, including rain streaks and background information, which can be used to selectively whiten (remove) the background information for storing only rain in memory networks.
		Note that MOSS~\cite{Huang2020memory}, which uses a memory network for de-raining, often fails to remove rain streaks because the memory features are mixed with background and rain streaks, but the BSW loss efficiently removes the background to improve the discriminative power of the rain steak-relevant features.
		
		Specifically, we extract two covariance matrices by inferring from the queries, extracted from different input rain images $(I_{w,n}, I_{v,n})$, and compute the variance matrix from the differences between two different covariance matrices. We hypothesize that the variance matrix $\mathbf{V}$ implies the sensitivity of the corresponding covariance to the rain streaks. In other words, the covariance elements with high variance values encode different rain streaks, such as density, orientation, and intensity. We define the variance matrix $\mathbf{V} \in \mathbb{R}^{C \times C}$ as $\mathbf{V}=\boldsymbol{\sigma}^2$ from the mean $\boldsymbol{\mu}$ and variance $\boldsymbol{\sigma^2}$ for each element from two different covariance matrices as follows:
		\begin{equation}
			\boldsymbol{\mu}  = \frac{1}{2}(\boldsymbol{{\Sigma_s}} ({\mathbf{q}_{v,n}})  + \boldsymbol{{\Sigma_s}} ({\mathbf{q}_{w,n}}) ),
		\end{equation}
		\begin{equation}
			{\boldsymbol{\sigma}^2} = \frac{1}{2}({(\boldsymbol{{\Sigma_s}} {({\bf{q}}_{v,n}) - \boldsymbol{\mu} } )^2} + {(\boldsymbol{{\Sigma_s}} {({\bf{q}}_{w,n}) - \boldsymbol{\mu} } )^2}),
		\end{equation}	
		where $\boldsymbol{{\Sigma_s}}(\cdot)$ extracts the covariance matrix of the intermediate feature map from each query. 
		As a result, $\mathbf{V}$ consists of elements of the variance of each covariance element across different queries extracted from the input rain images.
		
		
		We split $s$ into two groups: $G_{low}=\{c_{1},...,c_{l}\}$ with a low variance value and $G_{high}=\{c_{l+1},...,c_{h}\}$ with a high variance value. 
		We assume that the rain streaks are encoded in the covariance belonging to $G_{high}$, and consistent background information is encoded in the covariance belonging to $G_{low}$.
		Therefore, the background selective whitening loss selectively suppresses only the background information-encoded covariance.
		Let the mask matrix $\mathbf{M} \in \mathbb{R}^{C \times C}$ for the BSW loss be such that
		\begin{equation}
			{\bf{M}} = \left\{ {\begin{array}{*{20}{c}}
					{1,}&{{\rm{if }}{\mkern 1mu} {\bf{V}} \in {G_{low}}}\\
					{0,}&\mathrm{otherwise}
				\end{array}} \right..
			\end{equation}
			The BSW loss is defined as
			\begin{equation}
				{{\cal L}_w} = \sum\limits_{i}^{}{\left\| { {\boldsymbol{\Sigma _s}(i) \otimes {\bf{M}}(i)} } \right\|_1},
			\end{equation}
			where $\otimes$ means an element-wise multiplication.
			
			It may be observed that the memory network captured the rain streak in Fig.~\ref{fig:5} (b)--(d).
			This shows that the vanilla memory network could not effectively discriminate the region between the background and rain streaks, which often limits the de-raining models to remove rain streaks.
			However, with our background selective loss, as shown in Fig.~\ref{fig:5} (f)--(h), the network clearly captures the rain streaks in the memory network and enables the effective removal of rain streaks in the real rain image when comparing the highlighted boxes, as shown in Fig.~\ref{fig:5} (a) and (e).
			Note that because the proposed BSW loss depends on the the paired rain images containing the same background but time-varying rain streaks, this loss is well-tailored to time-lapse data.
			

			\subsubsection{Total Loss}
			For training our network, the total loss function consists of the aforementioned loss functions. 
			These terms are balanced by the weights $\lambda_{b}$, $\lambda_{s}$, $\lambda_{c}$, and $\lambda_{w}$ as
			\begin{equation}~\label{eq:14}
				{{\mathcal{L}}}=\lambda_{b}{{\mathcal{L}}_{b}}+\lambda_{s}{{\mathcal{L}}_{s}}+\lambda_{c}{{\mathcal{L}}_{c}}+\lambda_{w}{{\mathcal{L}}_{w}}.
			\end{equation}

			\begin{figure*}[!]
				\centering
				\renewcommand{\thesubfigure}{}
				\subfigure{\includegraphics[width=0.19\textwidth,height=0.1\textheight]{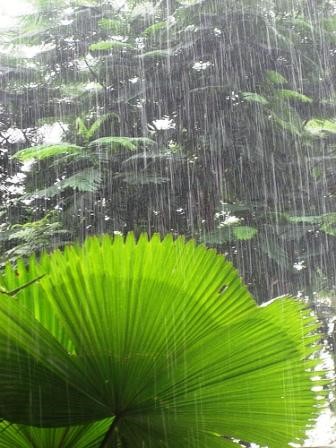}}
				\subfigure{\includegraphics[width=0.19\textwidth,height=0.1\textheight]{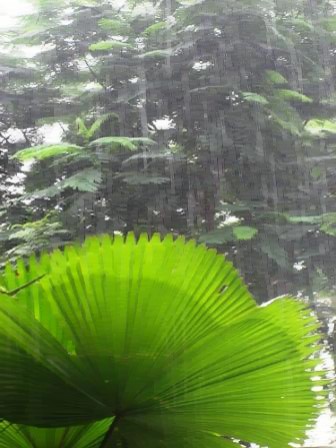}}
				\subfigure{\includegraphics[width=0.19\textwidth,height=0.1\textheight]{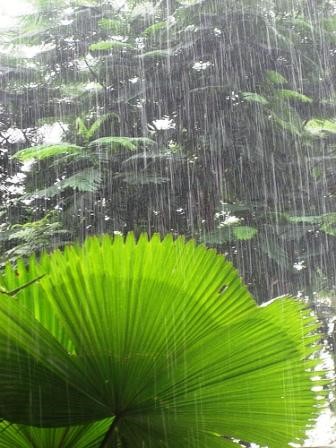}}
				\subfigure{\includegraphics[width=0.19\textwidth,height=0.1\textheight]{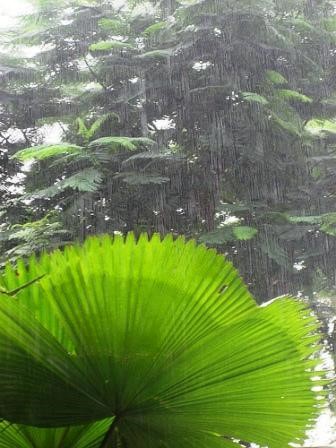}}
				\subfigure{\includegraphics[width=0.19\textwidth,height=0.1\textheight]{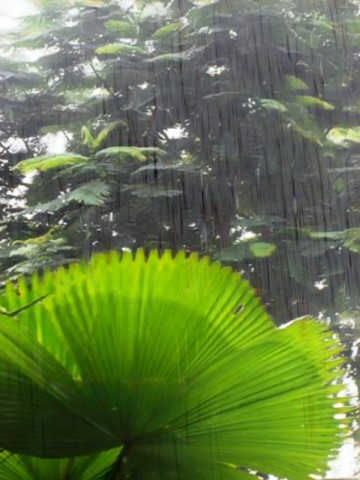}}
				\\ 
				\vspace{-8pt}
				\subfigure{\includegraphics[width=0.19\textwidth,height=0.09\textheight]{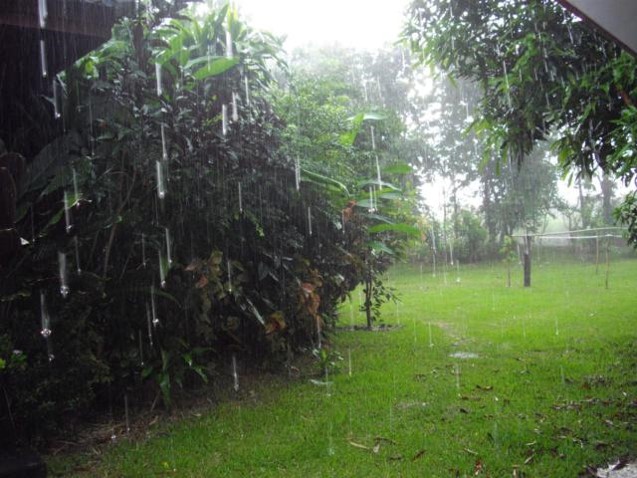}}
				\subfigure{\includegraphics[width=0.19\textwidth,height=0.09\textheight]{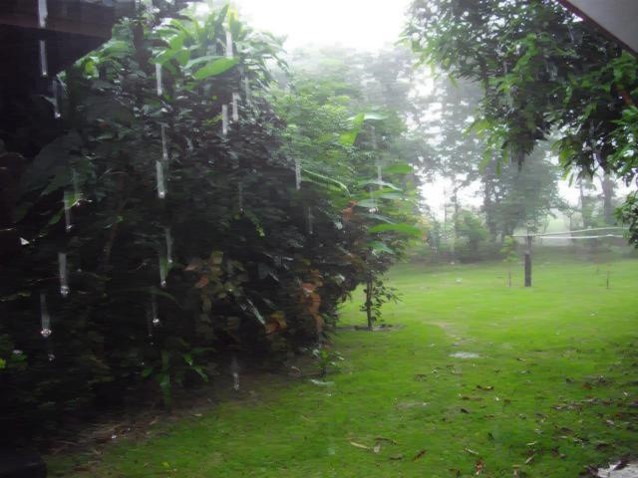}}
				\subfigure{\includegraphics[width=0.19\textwidth,height=0.09\textheight]{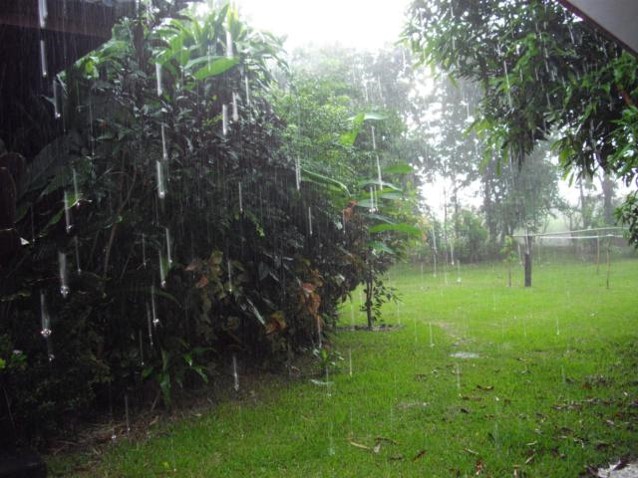}}
				\subfigure{\includegraphics[width=0.19\textwidth,height=0.09\textheight]{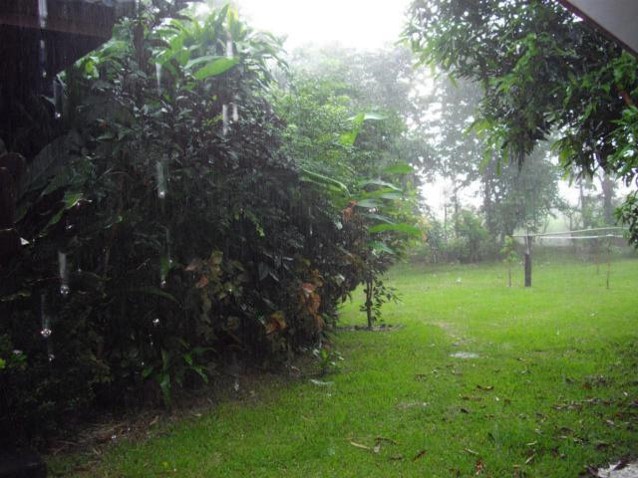}}
				\subfigure{\includegraphics[width=0.19\textwidth,height=0.09\textheight]{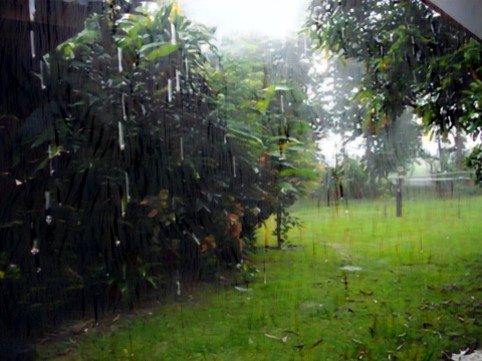}}
				\\ \vspace{-8pt}
				\subfigure[(a) Input]{\includegraphics[width=0.19\textwidth,height=0.08\textheight]{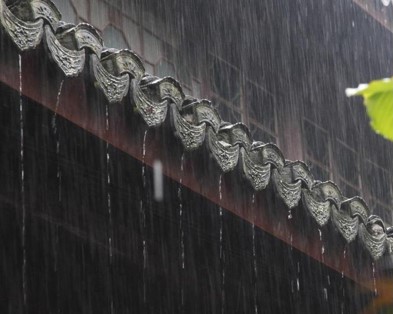}}
				\subfigure[(b) JCAS~\cite{gu2017joint}]{\includegraphics[width=0.19\textwidth,height=0.08\textheight]{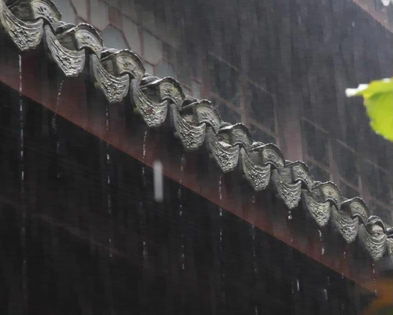}}
				\subfigure[(c) DID~\cite{zhang2018density}]{\includegraphics[width=0.19\textwidth,height=0.08\textheight]{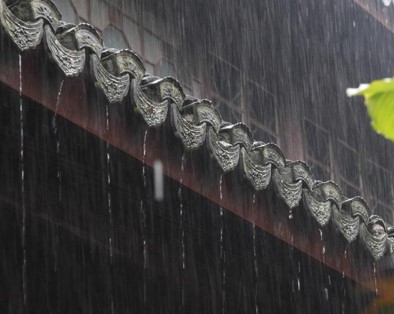}}
				\subfigure[(d) JORDER-E~\cite{yang2019joint}]{\includegraphics[width=0.19\textwidth,height=0.08\textheight]{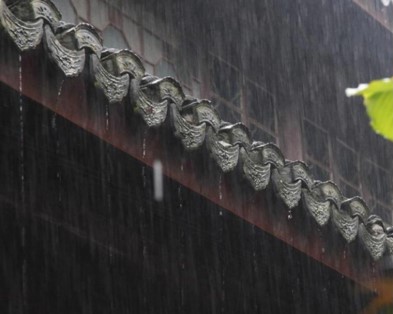}}	
				\subfigure[(e) SIRR~\cite{SIRR}]{\includegraphics[width=0.19\textwidth,height=0.08\textheight]{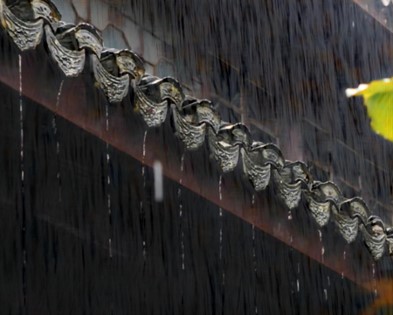}}
				\\ \vspace{-8pt}
				\subfigure{\includegraphics[width=0.19\textwidth,height=0.1\textheight]{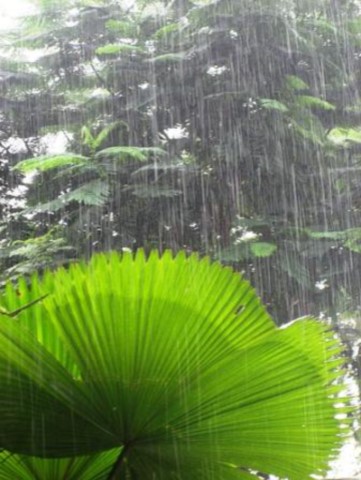}}
				\subfigure{\includegraphics[width=0.19\textwidth,height=0.1\textheight]{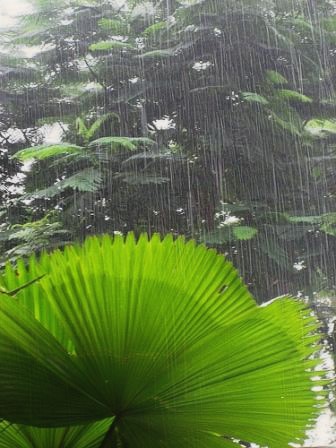}}
				\subfigure{\includegraphics[width=0.19\textwidth,height=0.1\textheight]{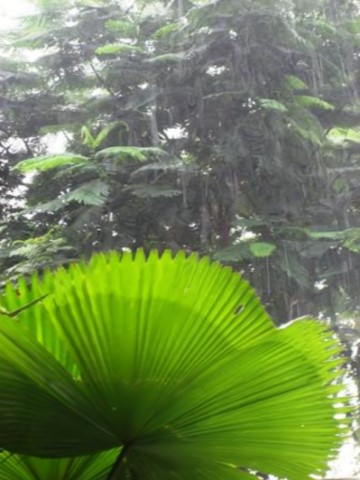}}
				\subfigure{\includegraphics[width=0.19\textwidth,height=0.1\textheight]{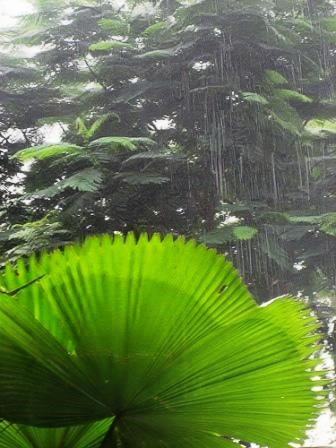}}
				\subfigure{\includegraphics[width=0.19\textwidth,height=0.1\textheight]{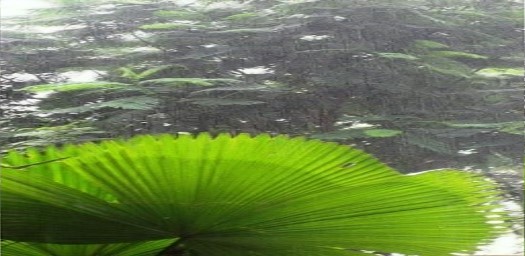}}	
				\\ \vspace{-8pt}
				\subfigure{\includegraphics[width=0.19\textwidth,height=0.09\textheight]{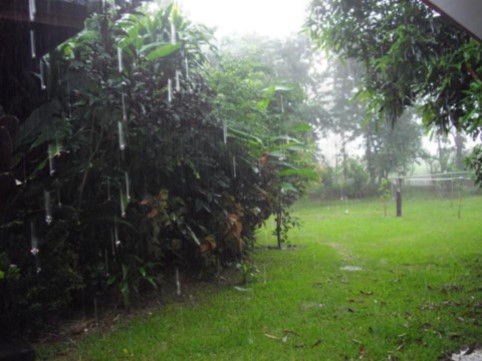}}
				\subfigure{\includegraphics[width=0.19\textwidth,height=0.09\textheight]{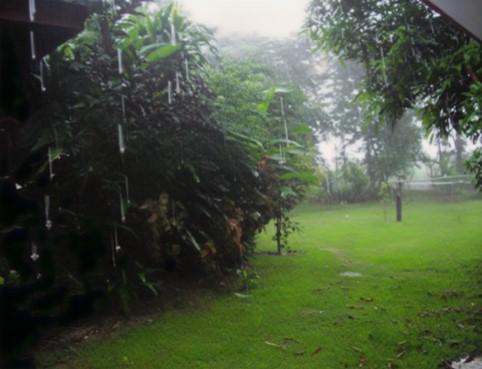}}
				\subfigure{\includegraphics[width=0.19\textwidth,height=0.09\textheight]{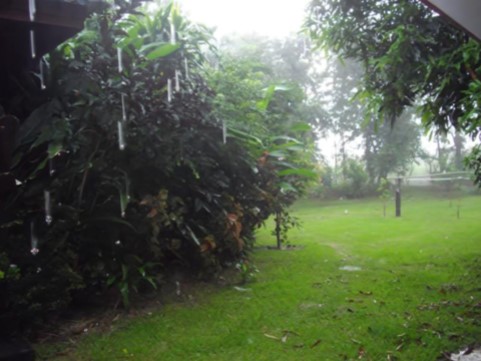}}
				\subfigure{\includegraphics[width=0.19\textwidth,height=0.09\textheight]{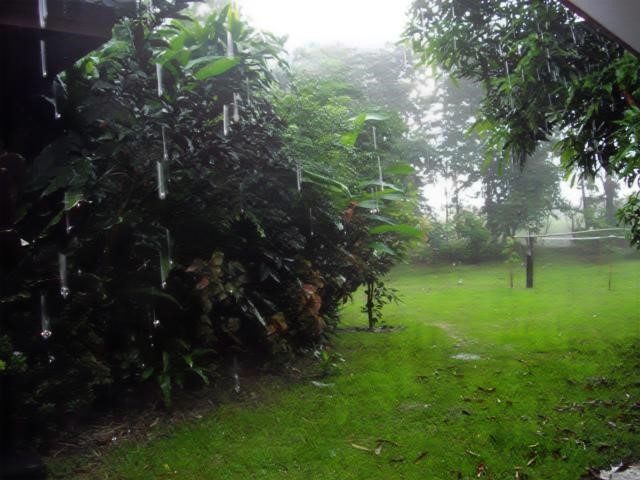}}
				\subfigure{\includegraphics[width=0.19\textwidth,height=0.09\textheight]{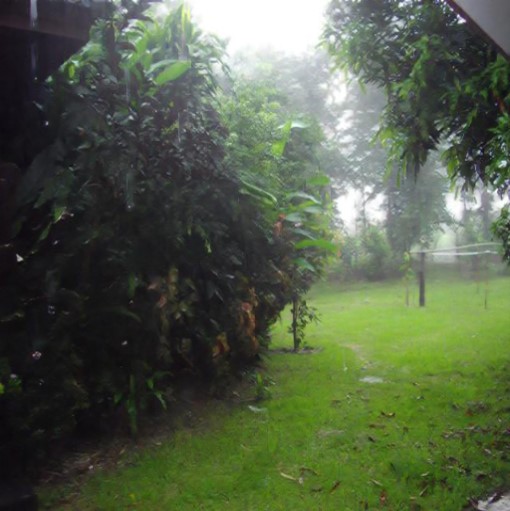}}
				\\ \vspace{-8pt}
				\subfigure[(f) RCDNet~\cite{wang2020model}]{\includegraphics[width=0.19\textwidth,height=0.08\textheight]{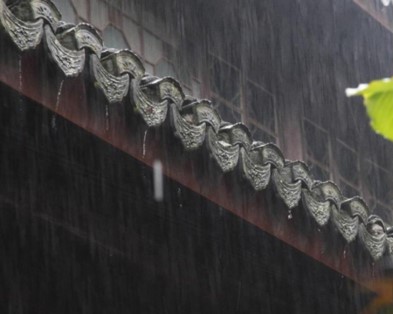}}
				\subfigure[(g) Syn2Real~\cite{yasarla2020syn2real}]{\includegraphics[width=0.19\textwidth,height=0.08\textheight]{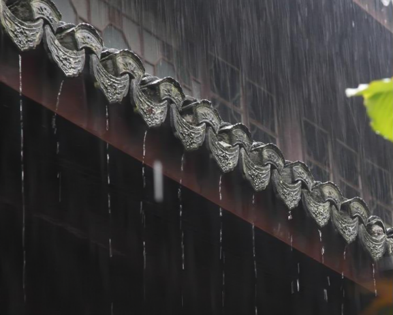}}
				\subfigure[(h) MPRNet~\cite{zamir2021multi}]{\includegraphics[width=0.19\textwidth,height=0.08\textheight]{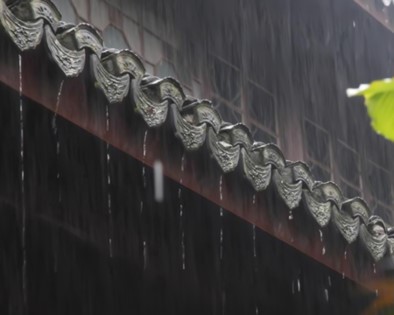}}
				\subfigure[(i) MOSS~\cite{Huang2020memory}]{\includegraphics[width=0.19\textwidth,height=0.08\textheight]{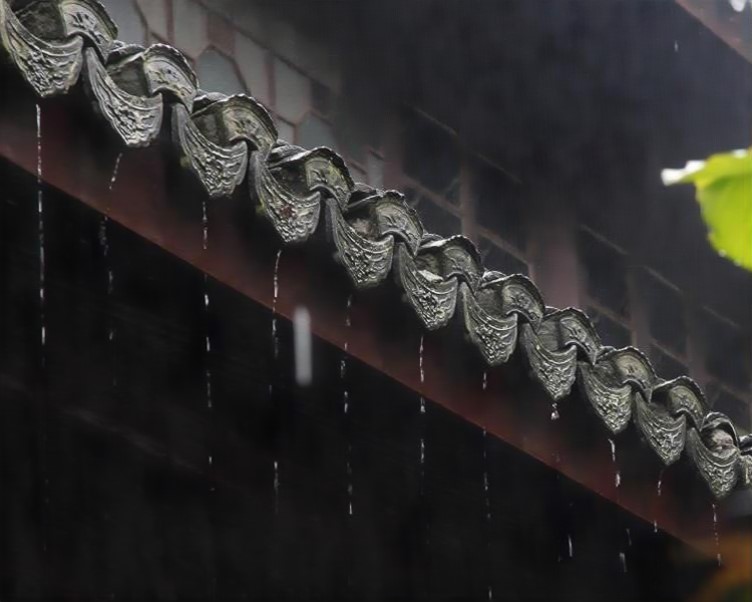}}
				\subfigure[(j) Ours]{\includegraphics[width=0.19\textwidth,height=0.08\textheight]{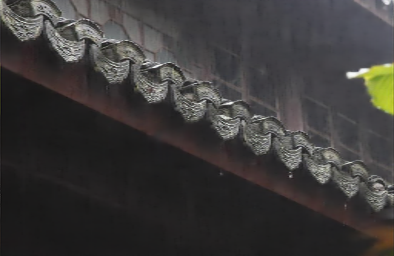}}
				\\			\vspace{-3pt}
				\caption{
					\textbf{Qualitative results on real rain images~\cite{yang2017deep,zhang2018density}.} (a) Input rain images and de-rained results of (b) JCAS~\cite{gu2017joint}, (c) DID~\cite{zhang2018density}, (d) JORDER~\cite{yang2017deep}, (e) SIRR~\cite{SIRR}, (f) RCDNet~\cite{wang2020model}, (g) Syn2Real~\cite{yasarla2020syn2real}, (h) MPRNet~\cite{zamir2021multi}, (i) MOSS~\cite{Huang2020memory} and (j) Ours.}
				\label{fig:real}	
				\vspace{-8pt}
			\end{figure*}

			\subsection{Implementation Details}
			The proposed networks were trained using the PyTorch~\cite{paszke2017automatic} library, with an Nvidia RTX TITAN GPU, which requires approximately 24 hours for training.
			We resized each input image to 256 $\times$ 256 and normalized it to the range [-1, 1]. Because our training data contained various real-world scenes, we do not use data augmentation such as flipping and rotating, as in~\cite{cho2020single}.
			We set the height $H$ and width $W$ of the query feature map, number of feature channels $C$, and memory items $M$ to 32, 32, 32, and 20, respectively.
			The memory items $\hat{\mathbf{p}}$ are randomly initialized.
			For the BSW loss, $l$ and $h$ are hyper-parameters, which are empirically set to $2$ and $4$, respectively.
			For the loss weighting parameters in Eq.~\ref{eq:14}, we empirically set $\lambda_{b}$ to be 1, $\lambda_{s}$ to  be 0.1, $\lambda_{c}$ to be 0.001, and $\lambda_{w}$.
			The model is trained using the Adam optimizer~\cite{kingma2014adam} with $\beta_{1}$= 0.9 and $\beta_{2}$ = 0.999, with a batch size of 16. 
			We set the initial learning rate to 2e-4 and decayed them using a cosine annealing method~\cite{loshchilov2016sgdr}.


			\section{Experiments}
			\subsection{Experimental Setup}
			In this section, we demonstrate the results of comprehensive experiments conducted to evaluate the performance of the proposed method, including comparisons with several state of the art single-image de-raining methods that include JCAS~\cite{gu2017joint}, DDN~\cite{fu2017removing}, DID~\cite{zhang2018density}, NLEDN~\cite{li2018non}, RESCAN~\cite{li2018recurrent}, PReNet~\cite{ren2019progressive}, JORDER-E~\cite{yang2019joint}, SIRR~\cite{SIRR}, SPANET~\cite{SPANet}, Cho \textit{et al.}~\cite{cho2020single}, Hu \textit{et al.}~\cite{hu2020single},  Syn2Real~\cite{yasarla2020syn2real}, RCDNet~\cite{wang2020model}, MPRNet~\cite{zamir2021multi}, and MOSS~\cite{Huang2020memory}. We used public code or pre-trained models provided by the authors to produce the de-raining results.
			
			In addition, we compared the performance of several video de-raining methods, including Garg \textit{et al.}~\cite{garg2007vision}, Kim \textit{et al.}~\cite{kim2015video}, Jiang \textit{et al.}~\cite{jiang2017novel}, Ren \textit{et al.}~\cite{ren2017video}, Wei \textit{et al.}~\cite{wei2017should}, Li \textit{et al.}~\cite{li2018video}, and Liu \textit{et al.}~\cite{liu2018erase}.
			Kim \textit{et al.}~\cite{kim2015video}\footnote{http://mcl.korea.ac.kr/deraining} and Li \textit{et al.}~\cite{li2018video}\footnote{https://github.com/MinghanLi/MS--CSC--Rain-Streak--Removal} provided the rain video dataset containing various forms of moving objects and background scenes, and different types of rain, varying from very light drizzle to heavy rain--storms and vertical rain to nearly horizontal rain.
			
			Using a single trained model, we evaluated both real and synthetic datasets.
			For quantitative evaluation, we measured the peak signal-to-noise ratio (PSNR) and the structure similarity index measure (SSIM).
			For evaluating on video dataset, we employed two additional metrics, namely VIF~\cite{sheikh2006image} and FSIM~\cite{zhang2011fsim}.
			We compared competing methods both qualitatively and quantitatively.

			\subsection{Datasets}
			\subsubsection{Training Dataset}
			To train our network, we used time-lapse benchmark provided
			by Cho \textit{et al.}~\cite{cho2020single}.
			\textbf{TimeLap} provided by Cho \textit{et al.}~\cite{cho2020single} consists of time-lapse sequences, where rain image pairs comprised 2 images sampled from 30 images from the 186 total scenes.
			For training and fair comparison, we mainly adopt the same experimental setups as Cho \textit{et al.}~\cite{cho2020single}. \textbf{RealDataset} provided by SPANET~\cite{SPANet} consists of 170 real rain videos captured by cell phone or collected from YouTube.
			\textbf{RealDataset} consists of 29,500 rain/clean image pairs image pairs using time-lapse data, which are split into 28,500 for training.
			Because our method requires time-lapse data without ground truth, we use only time-lapse data for training.
			

			\subsubsection{Test Dataset}
			We conducted experiments on real and synthetic datasets, respectively. 
			For the real dataset, we used the \textbf{RealDataset} to provide real rain images with realistic ground-truth background images generated by a semi-automatic algorithm~\cite{SPANet}, thus enabling the quantitative evaluations.
			This dataset consists of 1,000 pairs for testing at a resolution of 512 $\times$ 512, collected from 170 video sequences.
			Additionally, we obtained real-world rain images scraped from the Internet and previous studies~\cite{zhang2018density,SPANet,yang2019joint} and used them for qualitative evaluation only.
			Furthermore, because real rain images often contain \emph{fog}, we also collected real rain images with \emph{fog} from the Internet.
			
			For the synthetic dataset, we used three datasets provided by DDN~\cite{fu2017removing}, DID~\cite{zhang2018density}, and JORDER-E~\cite{yang2019joint}.
			\textbf{Rain14000} constructed by DDN~\cite{fu2017removing} provides 14,000 rain/clean image pairs synthesized from 1000 clean images with 14 kinds of different rain directions and scales.
			\textbf{Rain12000} constructed by DID~\cite{zhang2018density} provides 12,000 rain/clean image pairs containing rain with different orientations and scales, where the number of images with light, medium, and heavy rain is all 4,000, respectively.
			As pointed out in ~\cite{li2018non,cho2020single}, because the synthesized examples in Rain100H are inconsistent with real images, we used \textbf{Rain100}, which consists of 100 light rain images/clean image pairs for training and testing, respectively.

			\subsection{Single Image De-raining Results}
			\subsubsection{Results on Real-World Data}
			To evaluate the generalization ability of all competing methods and our method, we first conducted a qualitative evaluation using real rain images collected from the Internet provided by~\cite{yang2017deep,SPANet,SIRR}.
			Fig.~\ref{fig:real} shows the de-rained results that our method removes rain streaks well while preserving more detailed information such as background and texture information.
			In particular, ~\cite{zhang2018density,wang2020model} often fail to capture long and thin rain in Fig.~\ref{fig:real} (c) and (f).
			However, our method effectively handles various types of rain streaks owing to the advantage of being able to store real-world rain streaks with a memory network during training.

\begin{figure*}
				\centering
				\renewcommand{\thesubfigure}{}
				\subfigure[(a) Input]{\includegraphics[width=0.161\textwidth,height=0.1\textheight]{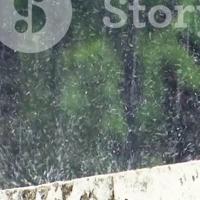}}
				\subfigure[(b) Ground truth]{\includegraphics[width=0.161\textwidth,height=0.1\textheight]{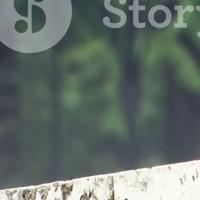}}
				\subfigure[(c) JCAS~\cite{gu2017joint}]{\includegraphics[width=0.161\textwidth,height=0.1\textheight]{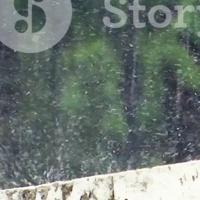}}
				\subfigure[(d) JORDER-E~\cite{yang2017deep}]{\includegraphics[width=0.161\textwidth,height=0.1\textheight]{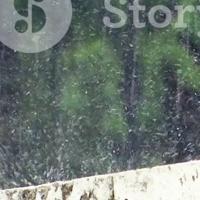}}
				\subfigure[(e) PReNet~\cite{ren2019progressive}]{\includegraphics[width=0.161\textwidth,height=0.1\textheight]{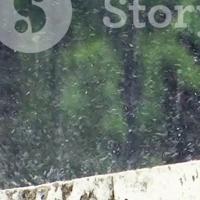}} 
				\subfigure[(f) SPANet~\cite{SPANet}]{\includegraphics[width=0.161\textwidth,height=0.1\textheight]{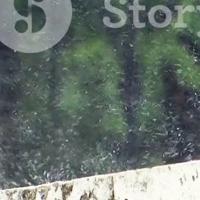}}\\ \vspace{-8pt}
				\subfigure[(g) Cho \textit{et al.}~\cite{cho2020single}]{\includegraphics[width=0.161\textwidth,height=0.1\textheight]{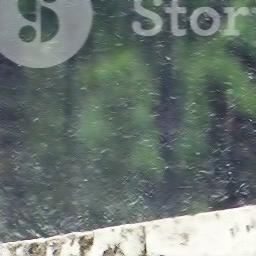}}
				\subfigure[(h) RCDNet~\cite{wang2020model}]{\includegraphics[width=0.161\textwidth,height=0.1\textheight]{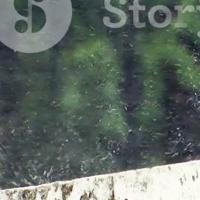}}
				\subfigure[(i) Syn2Real~\cite{yasarla2020syn2real}]{\includegraphics[width=0.161\textwidth,height=0.1\textheight]{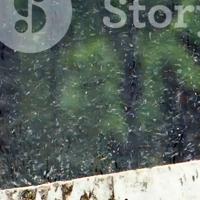}}
				\subfigure[(j) MPRNet~\cite{zamir2021multi}]{\includegraphics[width=0.161\textwidth,height=0.1\textheight]{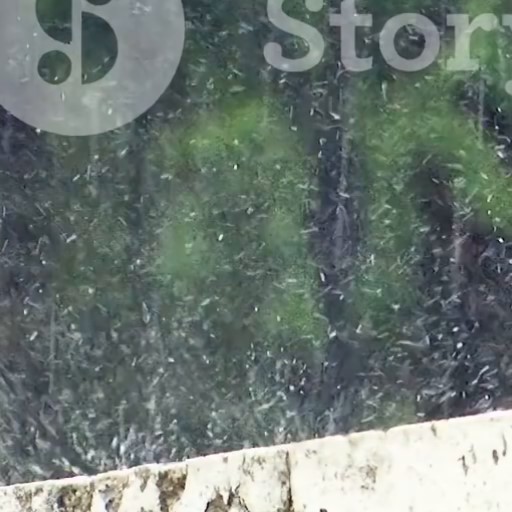}}
				\subfigure[(k) MOSS~\cite{Huang2020memory}]{\includegraphics[width=0.161\textwidth,height=0.1\textheight]{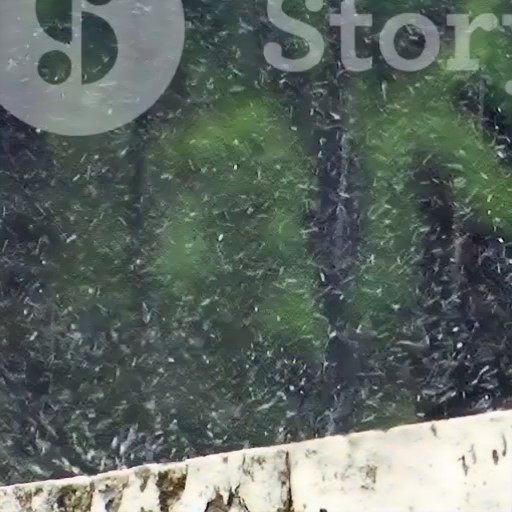}}
				\subfigure[(l) Ours]{\includegraphics[width=0.161\textwidth,height=0.1\textheight]{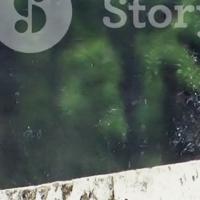}}\\ 			\vspace{-3pt}
				\caption{
					\textbf{Qualitative results on {RealDataset}~\cite{SPANet}.} 
					(a) Input rain images and (b) ground truth background images and obtained results of (c) JCAS~\cite{gu2017joint}, (d) JORDER-E~\cite{yang2019joint}, (e) PReNet~\cite{ren2019progressive}, (f) SPANet~\cite{SPANet}, (g) Cho \textit{et al.}~\cite{cho2020single}, (h) RCDNet~\cite{wang2020model}, (i) Syn2Real~\cite{yasarla2020syn2real}, (j) MPRNet~\cite{zamir2021multi}, (k) MOSS~\cite{Huang2020memory}  and (l) Ours.}
				\label{fig:SPANet_Compare}
				\vspace{-10pt}
			\end{figure*}

			Furthermore, we conducted experiments on publically available real-world dataset provided by SPANET~\cite{SPANet}.
			As shown in Fig.~\ref{fig:SPANet_Compare}, traditional hand-crafted method, \textit{that is}, JCAS~\cite{gu2017joint}, encountered difficulty in removing rain artifacts.
			Although CNN--based methods~\cite{SPANet,wang2020model} remove rain streaks better than hand-crafted method, they still suffer from the dot-patterned rain streaks.
			In contrast, our method preserves background details better and effectively removes various rain streaks including dot-patterned and long-shaped rain streaks. 
			In addition, we conducted a quantitative evaluation using the \textbf{RealDataset}~\cite{SPANet} in the fourth column of Table~\ref{tab:3}. 
			Interestingly, the hand-crafted methods~\cite{luo2015removing,li2016rain,gu2017joint} outperformed the CNN--based method~\cite{zhang2018density}.
			This shows the limitation of the fully supervised learning paradigm because such method tends to fail when dealing with conditions of real rain streaks that have never been encountered during training.
			Our model achieved the best results by leveraging the real-world time-lapse data without ground truth.

				\begin{table*}
					\centering
					\caption{
						\textbf{Quantitative comparison of single image de-raining using synthetic and real-world datasets.} 
						GT and T-L denote using paired ground truth data and time-lapse data, respectively. 
						$\dagger$, $\ddagger$, and $\sharp$ indicate that the methods require additional supervised cues such as, binary mask map, rain density level, and attention maps, respectively. 
						The best result is shown in bold, and the second--best is underlined. The higher the PSNR and SSIM is, the better.
					}
					\vspace{-5pt}
					\renewcommand{\tabcolsep}{4.1mm}
					\begin{tabular}{l|c|c|cc|cc|cc|cc}
						\toprule 
						\multirow{3}{*}{Benchmark} &  \multirow{3}{*}{GT} &  \multirow{3}{*}{T-L} & \multicolumn{6}{c|}{Synthetic Dataset}  & \multicolumn{2}{c}{Real-world Dataset} \\ \cline{4-11}
						&  &  & \multicolumn{2}{c|}{\textbf{Rain14000}}  & \multicolumn{2}{c|}{\textbf{Rain12000}} & \multicolumn{2}{c|}{\textbf{Rain100}}  & \multicolumn{2}{c}{{\textbf{RealDataset}}} \\ \cline{4-11}
						&   &  &  PSNR & SSIM &  PSNR & SSIM  &  PSNR & SSIM   &  PSNR & SSIM   \\ 
						\midrule
						DSC~\cite{luo2015removing} & No      & No         & 27.88 & 0.839 &    24.24  & 0.828 &    27.16  & 0.866   &   34.15   &  0.927     \\
						GMM~\cite{li2016rain}      & No   & No          & 27.78 & 0.859 &   25.81 & 0.834 &   28.66   & 0.865  &  34.30    &  0.943    \\
						JCAS~\cite{gu2017joint}      & No     & No    & 26.20 & 0.847  &  25.16 & 0.851 &   31.42   & 0.917  &   34.95   & 0.945     \\
						DDN~\cite{fu2017removing}  & Yes    & No         & 28.45 & 0.889 & 30.97 & 0.912 &   34.68   & 0.967   &  36.16   & 0.946    \\
						DID~\cite{zhang2018density}$\dagger$& Yes   & No      & {26.17} & {0.887} &  31.30 & 0.921 & 35.40 &  0.962   &  28.96    & 0.941      \\
						NLEDN~\cite{li2018non}& Yes     & No      & {29.79} & {0.897} & {33.16} & {0.919} & 36.57 &  0.975   &  40.12    &  0.984    \\
						PReNet~\cite{ren2019progressive}& Yes  & No       & {{32.55}} & {\textbf{{0.946}}} &  33.17 & 0.942 &   37.80   & {{0.981}}  &  40.16    &  0.982     \\
						SIRR~\cite{SIRR}& Yes   & No      & 28.44 & 0.889 & 30.57 & 0.910 &   34.75   & 0.969    &  35.31    &  0.941   \\
						JORDER-E~\cite{yang2019joint}$\ddagger$ & Yes   & No      & 32.00 & {{{0.935}}} & \underline{33.98} & {\textbf{0.950}} &   \underline{{38.59}}  & \underline{{0.983}}     &  {{40.78}}     &  0.981   \\
						Syn2Real~\cite{yasarla2020syn2real}  & Yes  & No & 29.23 & 0.898 &  30.90 & 0.878 &  36.09 & 0.967  &  37.87    &  0.965   \\ 
						RCDNet~\cite{wang2020model}  & Yes  & No  & 30.66 & 0.921 &  31.99 & 0.921 &  {\textbf{40.17}}  & {\textbf{0.988}}  &  \underline{{41.47}}   &  \underline{{0.983}}    \\
						\midrule
						SPANet~\cite{SPANet}$\sharp$& Yes   & Yes      & {29.85} & {0.912} &  33.04 & \underline{{0.949}} &   35.79   & 0.965  &  40.24    &  0.981    \\
						Cho {et al.}~\cite{cho2020single}  & No  & Yes  & \underline{{{33.73}}} & 0.941 &  {{33.25}} & 0.935 &    37.89  &{{0.980}}   &  38.54    &   {\textbf{0.989}}   \\
						\midrule
						Ours  & No & Yes  & {\textbf{34.02}} & \underline{{0.945}} & {\textbf{34.55}}& \underline{{{0.949}}} &  {{38.45}}& {{0.981}}  &  {\textbf{41.56}}    & \textbf{{0.989}} \\
						\bottomrule
					\end{tabular}
					\label{tab:3}\vspace{-10pt}
				\end{table*}
			
			
			\subsubsection{Results on Synthetic Data}	
			From the first to third row in Table~\ref{tab:3}, we show the quantitative results of recent de-raining methods when trained on synthetic data including \textbf{Rain14000}, \textbf{Rain100}, and \textbf{Rain12000}.
			We point out that, excluding DID~\cite{zhang2018density} in \textbf{Rain14000}, CNN--based methods~\cite{cho2020single,SPANet,wang2020model,yasarla2020syn2real,fu2017removing,li2018non,ren2019progressive,SIRR,yang2019joint} outperform the hand-crafted methods methods on a synthetic dataset.
			In addition, the results show that the existing CNN--based methods exhibited significant performance differences depending on the test set.
			For example, RCDNet~\cite{wang2020model} has the best performance in the \textbf{Rain100} dataset, but the fifth--best performance in \textbf{Rain14000}.
			These results show that supervised learning--based methods vary greatly in performance depending on the characteristics of the synthetic datasets.
			In contrast, because our method does not rely on supervised learning, it achieved a comparatively consistent performance in various synthetic datasets, showing better generalization ability.
			The proposed approach achieved the best performance on \textbf{Rain14000} and \textbf{Rain12000} in terms of PSNR and showed a high generalization capability on other synthetic datasets, including \textbf{Rain100}.
			We believe that models trained with a memory network and BSW loss can alleviate the domain shift of the existing methods.
			
\begin{figure*}
								\centering
								\subfigure{\includegraphics[width=1\textwidth,height=0.37\textheight]{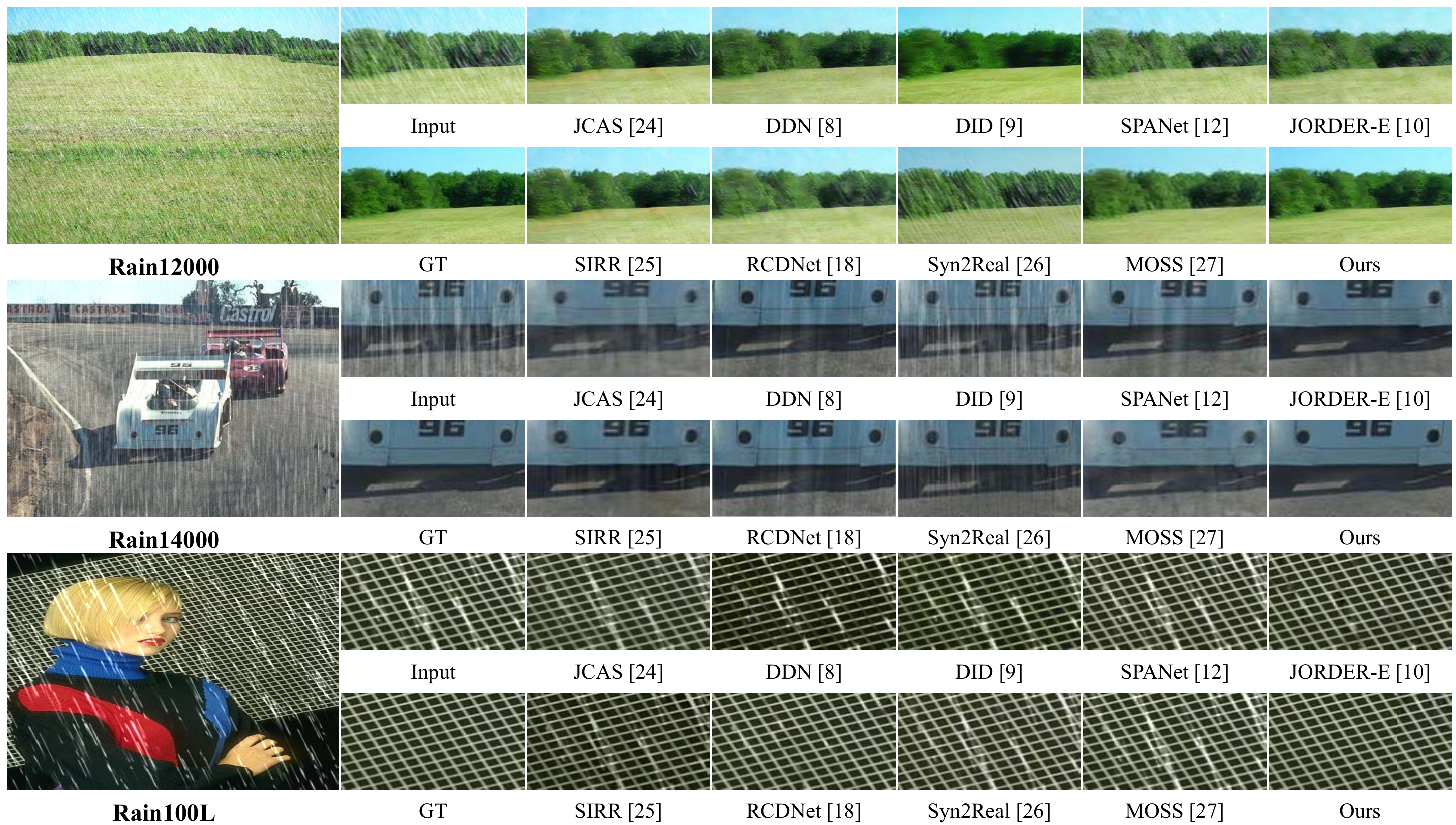}}
								\vspace{-15pt}
								\caption{
									\textbf{Qualitative results on different synthetic datasets including {Rain14000}, \textbf{Rain12000}, and {Rain100}.}}
								\label{fig:synthetic}
								\vspace{-8pt}
							\end{figure*}

			We further compare the visualization results of our method with those of the state of the art methods in Fig.~\ref{fig:synthetic}.
			Our method, designed to focus on real-world rain streaks, can also effectively handle synthetic rain streaks and achieve better results.
			Our method preserves details while effectively removing rain, demonstrating that our method could discriminate the rain streaks and background scene better than the existing methods.

			
			\subsubsection{Analysis on Rain Image with Fog}
\begin{figure*}
						\centering
						\renewcommand{\thesubfigure}{}
						\subfigure[Input]{\includegraphics[width=0.165\textwidth,height=0.07\textheight]{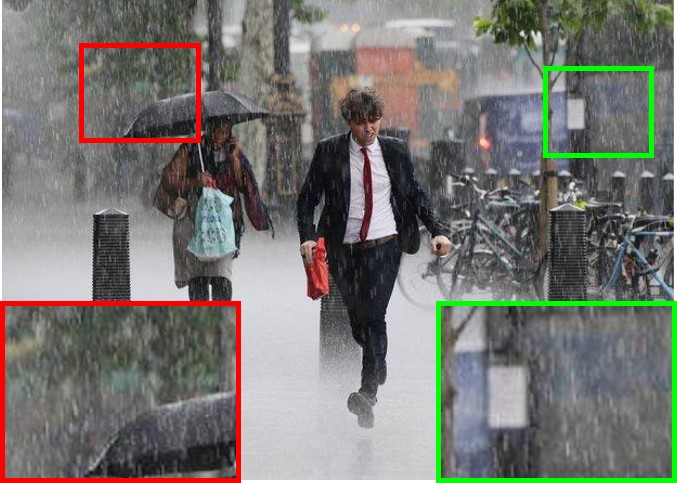}}
						\subfigure[Hu \textit{et al.}~\cite{hu2020single}]{\includegraphics[width=0.161\textwidth,height=0.07\textheight]{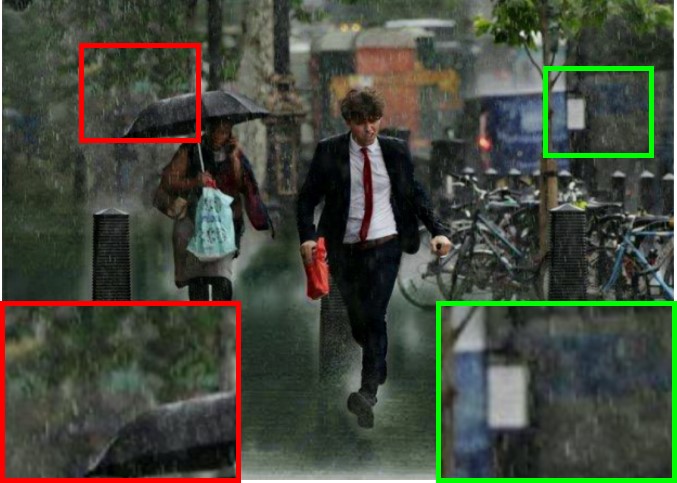}}
						\subfigure[SPANet~\cite{SPANet}]{\includegraphics[width=0.161\textwidth,height=0.07\textheight]{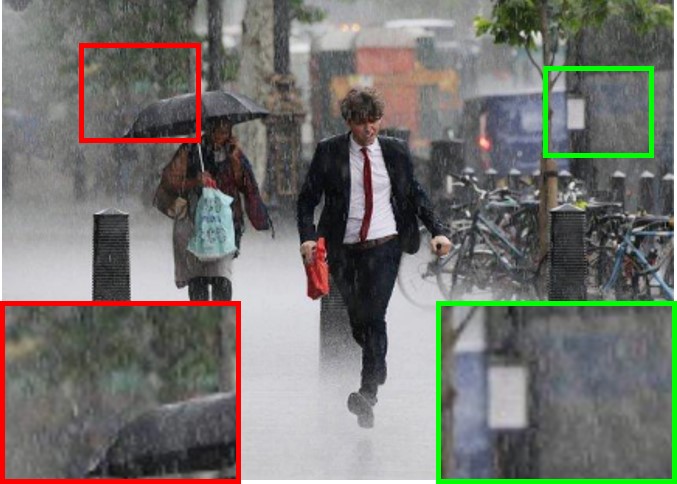}}
						\subfigure[SIRR~\cite{SIRR}]{\includegraphics[width=0.161\textwidth,height=0.07\textheight]{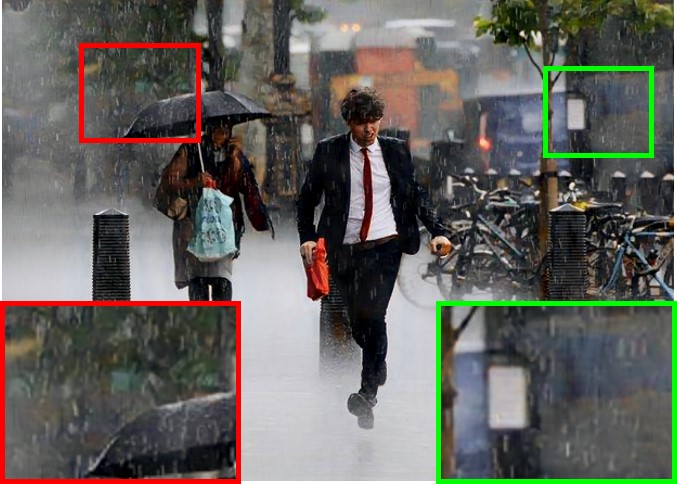}}
						\subfigure[Syn2Real~\cite{yasarla2020syn2real}]{\includegraphics[width=0.161\textwidth,height=0.07\textheight]{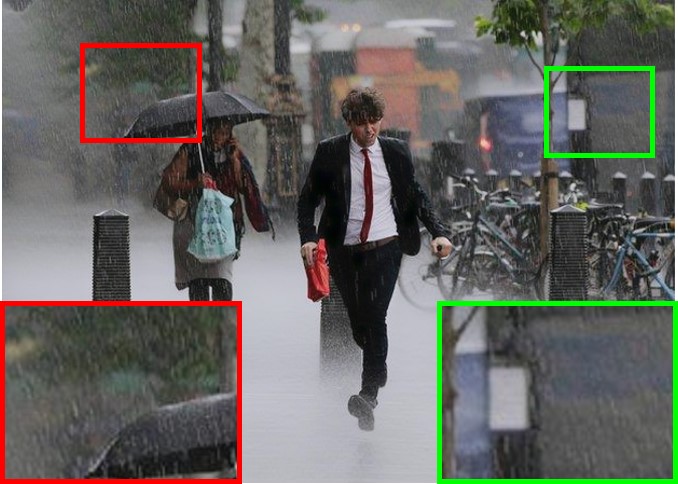}}
						\subfigure[Ours]{\includegraphics[width=0.161\textwidth,height=0.07\textheight]{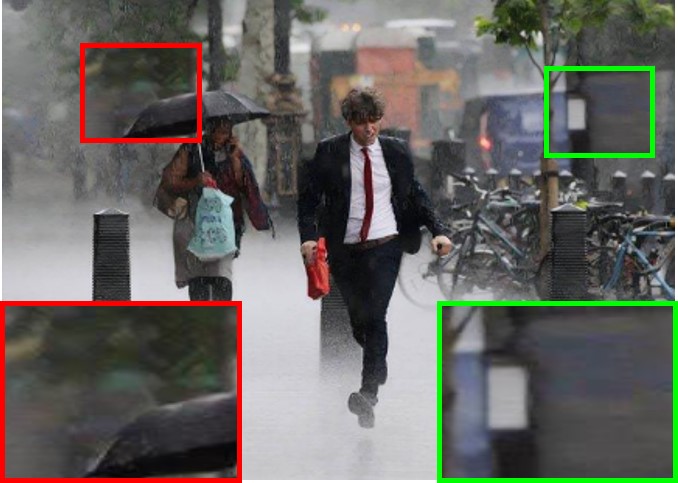}}\\ \vspace{-8pt}
						\subfigure[Input]{\includegraphics[width=0.161\textwidth,height=0.07\textheight]{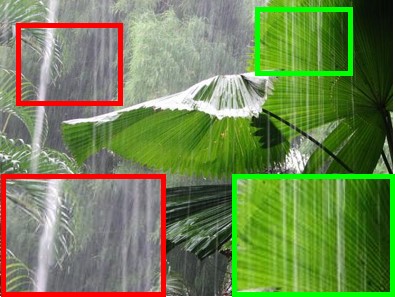}}
						\subfigure[Hu \textit{et al.}~\cite{hu2020single}]{\includegraphics[width=0.161\textwidth,height=0.07\textheight]{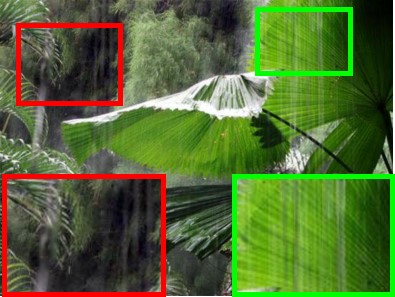}}
						\subfigure[JORDER-E~\cite{yang2019joint}]{\includegraphics[width=0.161\textwidth,height=0.07\textheight]{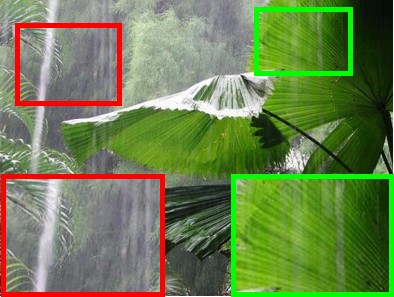}}
						\subfigure[RCDNet~\cite{wang2020model}]{\includegraphics[width=0.161\textwidth,height=0.07\textheight]{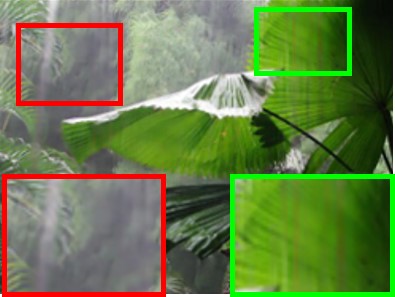}}
						\subfigure[MOSS~\cite{Huang2020memory}]{\includegraphics[width=0.161\textwidth,height=0.07\textheight]{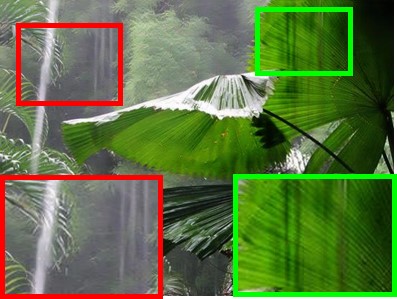}}
						\subfigure[Ours]{\includegraphics[width=0.161\textwidth,height=0.07\textheight]{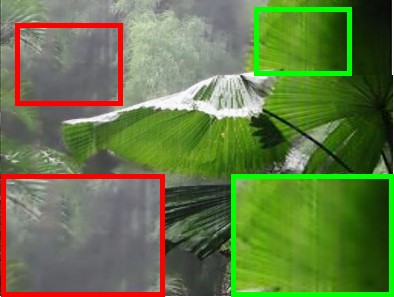}}
						\vspace{-13pt}
						\caption{
							\textbf{Results of the proposed method and the state of the art methods on real rain images accompanied by fog.
							}
						}
						\vspace{-10pt}
						\label{fig:fog}
					\end{figure*}

			For further analysis, we analyzed whether the proposed method could estimate rain streaks even with fog.
			We also compared with state of the art methods including Hu \textit{et al.}~\cite{hu2020single}, which proposed a depth-guided attention mechanism to remove rain and fog simultaneously. 
			Fig.~\ref{fig:fog} illustrated the comparison results. While existing methods are difficult to remove rain streaks with fog, our method is successful in removing the rain streaks even with fog. 
			Hu \textit{et al.}~\cite{hu2020single} tended to leave rain streaks. In addition, they often show the artifacts observed in the object boundaries in the first row in Fig.~\ref{fig:fog}.
			Alhough Hu \textit{et al.}~\cite{hu2020single} can handle fog, it is still challenging to remove rain streaks.
			In contrast, our method can handle rain streaks affected by fog in the real rain images because the proposed memory-based method with BSW loss captured real-world rain streaks well.
			
			
			
		\begin{table}
			\centering
			\caption{
				\textbf{Quantitative comparison of video de-raining methods on a video dataset~\cite{li2018video}.}}
			\vspace{-5pt}
			{   \begin{tabular}[b]{c|cccc}
					\toprule
					Metrics	& PSNR ($\uparrow$) & VIF ($\uparrow$) & FSIM ($\uparrow$) & SSIM ($\uparrow$) \\ \hline
					Garg \textit{et al.}~\cite{garg2007vision} & 24.15 & 0.611 & 0.970 & 0.911 \\
					Kim \textit{et al.}~\cite{kim2015video} & 22.39 & 0.526 & 0.960 & 0.886 \\ 
					Jiang \textit{et al.}~\cite{jiang2017novel} & 24.32 & 0.713 & 0.932 & 0.938 \\ 
					Ren \textit{et al.}~\cite{ren2017video} & 23.52 & 0.681 & 0.966 & 0.927 \\ 
					Wei \textit{et al.}~\cite{wei2017should} & 24.47 & 0.779 & 0.966 & 0.951 \\ 
					Li \textit{et al.}~\cite{li2018video} & 25.37 & 0.790 & 0.980 & 0.957 \\ 
					Liu \textit{et al.}~\cite{liu2018erase} & 22.19 & 0.555 & 0.980 & 0.895 \\ \hline
					Ours & 24.21 & 0.722 & 0.967 & 0.913  \\
					\bottomrule
				\end{tabular}}
				\vspace{-10pt}
				\label{tab:video}
			\end{table}

					\begin{figure}
						\centering
						\renewcommand{\thesubfigure}{}
						\subfigure[Input]{\includegraphics[width=0.15\textwidth,height=0.1\textheight]{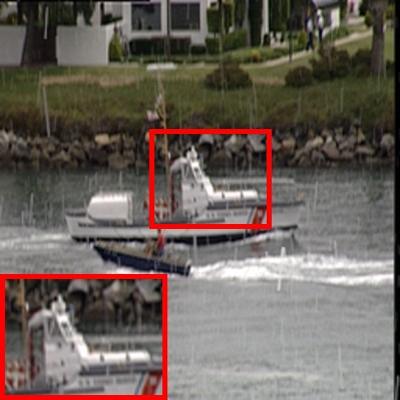}}
						\subfigure[Kim \textit{et al.}~\cite{kim2015video}]{\includegraphics[width=0.15\textwidth,height=0.1\textheight]{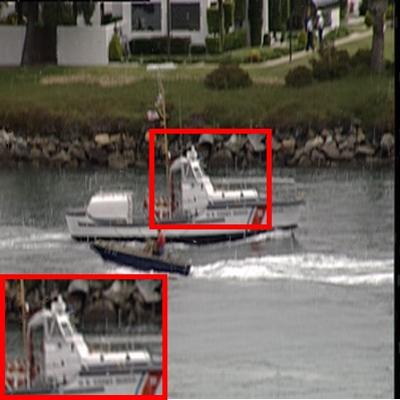}}
						\subfigure[Jiang \textit{et al.}~\cite{jiang2017novel}]{\includegraphics[width=0.15\textwidth,height=0.1\textheight]{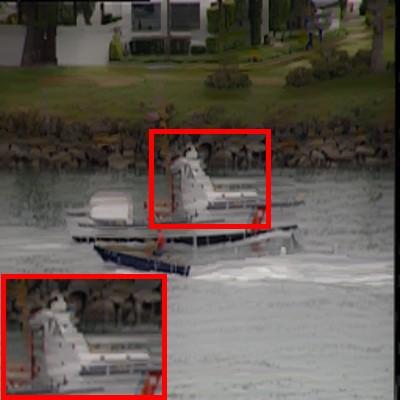}}
						\\ \vspace{-8pt}
						\subfigure[Chen \textit{et al.}~\cite{chen2018robust}]{\includegraphics[width=0.15\textwidth,height=0.1\textheight]{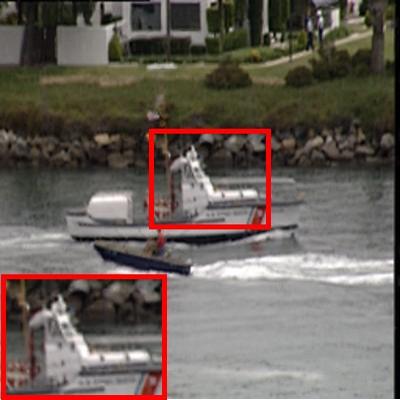}} 
						\subfigure[Li \textit{et al.}~\cite{li2018video}]{\includegraphics[width=0.15\textwidth,height=0.1\textheight]{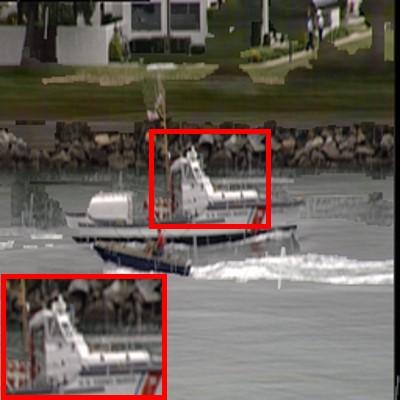}}
						\subfigure[Ours]{\includegraphics[width=0.15\textwidth,height=0.1\textheight]{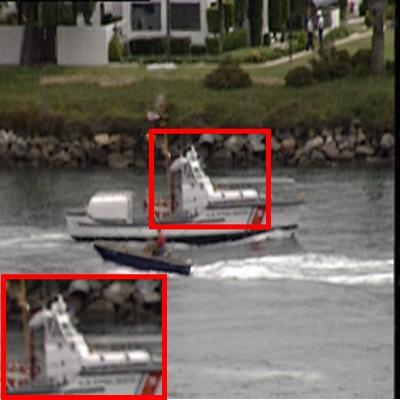}}
						\vspace{-5pt}
						\caption{
							\textbf{Visual comparison on a rain video dataset~\cite{li2018video}.}}
						\vspace{-10pt}
						\label{fig:Video}
					\end{figure}

				\subsection{Video De-raining Results}
				Although our method requires a single image at the testing time, it can perform well even for video data. We conducted experiments on videos containing various types of synthetic rain streaks, as shown in Fig.~\ref{fig:Video}.
				Kim \textit{et al.}~\cite{kim2015video}, and Li \textit{et al.}~\cite{li2018video} did not effectively remove rain. 
				Although the methods, proposed by Jiang \textit{et al.}~\cite{jiang2017novel} and Chen \textit{et al.}~\cite{chen2018robust}, exhibited a better removing rain streaks, the results showed a blurry artifact in the moving objects.
				While our method requires only a single image at inference time, our result removes rain effectively and preserves the details of the background and moving object.
				Table~\ref{tab:video} shows the quantitative comparisons with these competing methods and recent works, on the highway dataset.
				It should be noted that although all the abovementioned methods require multiple images at training and testing time, our method achieved fairly plausible results using single image at the testing time compared to existing video de-raining methods.

								\begin{table}
									\centering
									\caption{\textbf{Ablation study in terms of the network architecture.}
									}\vspace{-5pt}
									\renewcommand{\tabcolsep}{4mm}
									\scalebox{0.9}{
										\begin{tabular}[b]{c|cccc}
											\toprule
											& 	$\mathcal{B}_{a}$  & $\mathcal{B}_{b}$ & $\mathcal{B}_{c}$  & $\mathcal{B}_{d}$  \\  
											\midrule
											Encoder-Decoder   &  \cmark  & \cmark   &     &   \\
											Siamese   &     &    &   \cmark  & \cmark  \\
											Memory   &     & \cmark  &    & \cmark \\
											\hline
											PSNR   &  36.62   & 38.03   &  39.82   & 41.08  \\
											SSIM   &  0.965  &  0.972  &  0.982   & 0.985 \\
											\bottomrule
										\end{tabular}}
										\label{tab:mem}
										\vspace{-8pt}
									\end{table}

					\subsection{Ablation Study}				
					In this section, we conducted an extensive ablation study to verify the importance of each component of the proposed method on the \textbf{RealDataset} and real-world rain images.

						\subsubsection{Analysis of Memory Network}
						To verify the effectiveness of the memory network, we evaluated the different combinations in Table~\ref{tab:mem}. First, $\mathcal{B}_{a}$ is used as a baseline encoder-decoder without a memory module. 
						$\mathcal{B}_{b}$ adds a memory network on $\mathcal{B}_{a}$. These were trained using rain and time-averaged clean images~\footnote{Given the characteristics of time-lapse sequences taken from static scenes, simply averaging all frames over time can be used for pseudo ground truth data.	We averaged 30 frames to produce pseudo ground truth.} pairs using \textbf{TimeLap}. To train the $\mathcal{B}_{a}$ and $\mathcal{B}_{b}$, we used the standard ${\mathcal{L}}_{1}$ loss to measure the per-pixel reconstruction accuracy~\cite{SPANet}. $\mathcal{B}_{c}$ is a Siamese network based on an encoder--decoder without a memory network. $\mathcal{B}_{d}$ adds a memory network to $\mathcal{B}_{c}$. To train $\mathcal{B}_{a}$ and $\mathcal{B}_{b}$, we used the loss functions including ${\mathcal{L}}_{b}$, ${\mathcal{L}}_{c}$, and ${\mathcal{L}}_{s}$. The analysis of the proposed BSW loss will be described in the following section. Table~\ref{tab:mem} shows that the proposed Siamese networks exhibited improved the performance compared to a simple encoder--decoder. The model trained with the memory network achieved a substantial accuracy gain over the model without a memory network. Because the memory network enables the learning of various rain streaks, the model with the memory network achieves the best performance, which proves that the memory network is helpful in improving de-raining performance.

										\begin{table}
											\centering
											\caption{
												\textbf{Ablation study of loss functions.}}\vspace{-5pt}
											\renewcommand{\tabcolsep}{4mm}
											\scalebox{0.9}{
												\begin{tabular}[b]{cccc|cc}
													\toprule
													$\mathcal{L}_{s}$  & $\mathcal{L}_{b}$ & $\mathcal{L}_{c}$  & $\mathcal{L}_{w}$ & PSNR  & SSIM \\  \cline{1-6}
													\midrule			
													\cmark     &    \cmark   &   &       & 39.92 & 0.981 \\		
													\cmark     &    \cmark   &   &   \cmark    & 40.35 & 0.982 \\		
													&    \cmark  &  \cmark   &     & 40.01 & 0.982 \\
													&    \cmark  &  \cmark   &  \cmark  & 40.57 & 0.984 \\	
													\cmark     &    \cmark   &  \cmark  &     & 41.08 & 0.985 \\
													\cmark     &    \cmark    &  \cmark  & \cmark   & 41.56 & 0.988 \\
													\bottomrule
												\end{tabular}}
												\label{tab:abla}
												\vspace{-8pt}
											\end{table}

											\begin{figure}
												\centering
												\renewcommand{\thesubfigure}{}
												\subfigure[(a) without BSW loss ]{\includegraphics[width=0.22\textwidth,height=0.12\textheight]{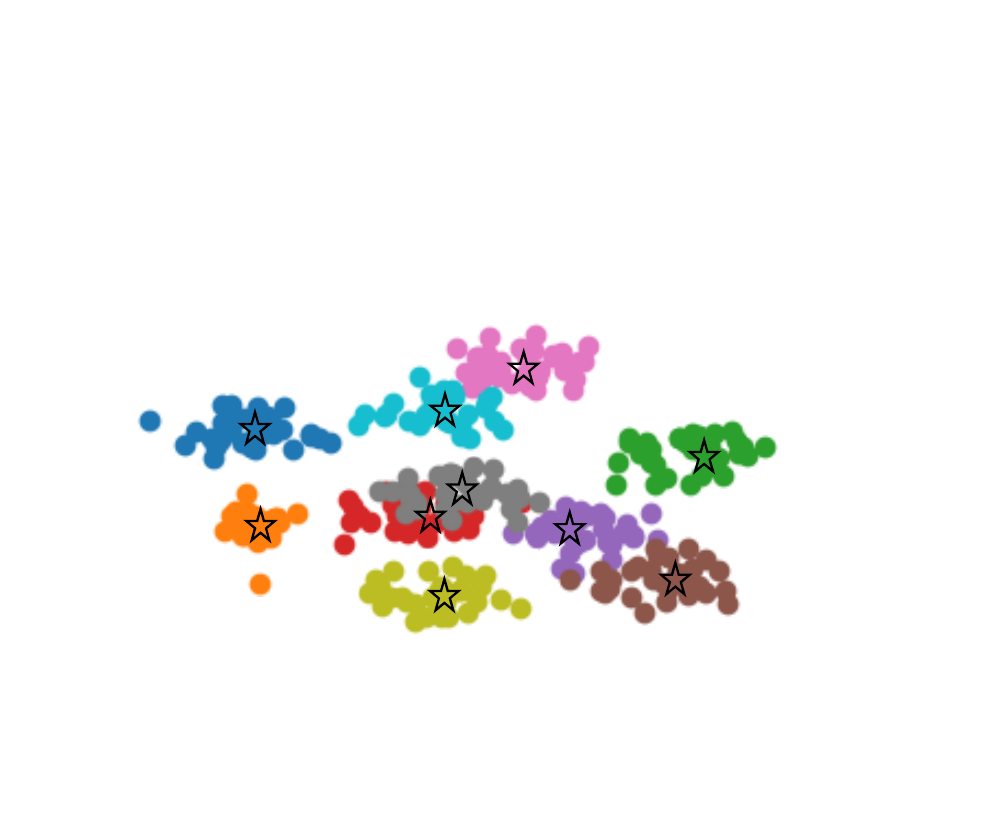}}
												\subfigure[(b) with BSW loss]{\includegraphics[width=0.22\textwidth,height=0.12\textheight]{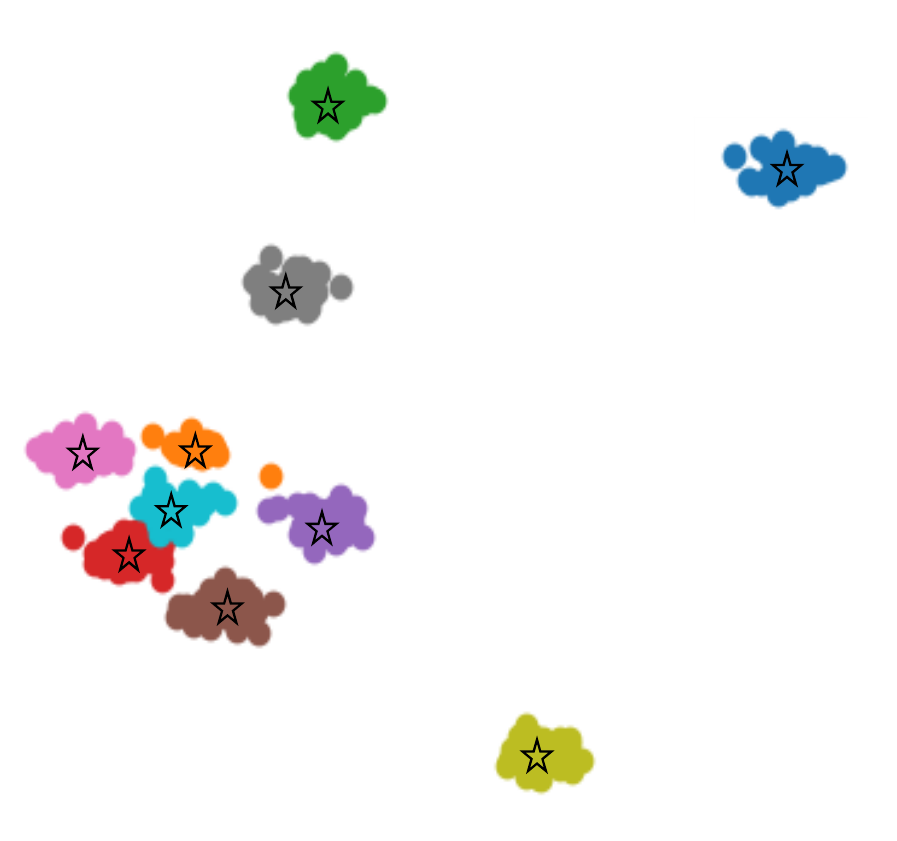}}
												\\ \vspace{-3pt}
												\caption{
													\textbf{Visualization of query features and memory items trained with/without BSW loss with t-SNE~\cite{van2014accelerating}.}
													For this visualization, we randomly sampled query features from real rain image. 
													The features and memory items are shown in points and stars, respectively. 
													The points with the same color are mapped to the same item. 
													The BSW loss enabled the separation of the items, recording the diverse prototypes of rain streaks.
													Owing to the BSW loss, the features were highly discriminative and similar rain streaks were clustered well (best viewed in color).}
												\label{fig:tsne}
												\vspace{-8pt}
											\end{figure}

						\begin{figure}
							\centering
							\renewcommand{\thesubfigure}{}
							\subfigure[(a)]{\includegraphics[width=0.24\textwidth,height=0.09\textheight]{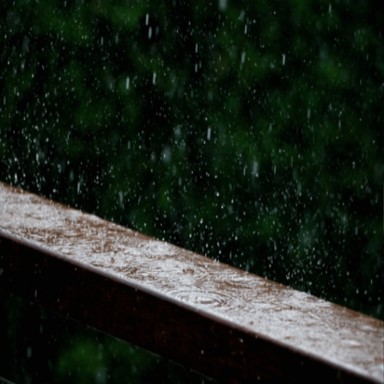}}
							\subfigure[(b)]{\includegraphics[width=0.24\textwidth,height=0.09\textheight]{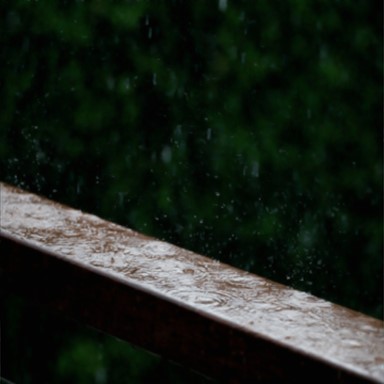}}\\ \vspace{-8pt}
							\subfigure[(c)]{\includegraphics[width=0.24\textwidth,height=0.09\textheight]{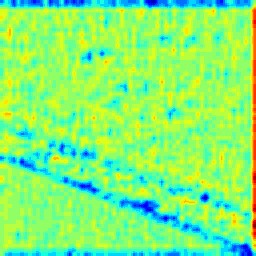}}
							\subfigure[(d)]{\includegraphics[width=0.24\textwidth,height=0.09\textheight]{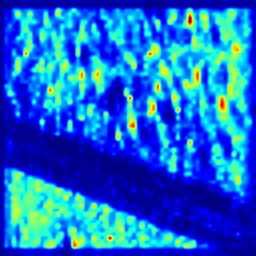}}\\
							\vspace{-3pt}
							\caption{
								\textbf{
									Visualization of de-raining result and memory features trained with/without background selective loss.} (a) Input rain image, (b) de-raining result, visualized feature maps from the decoder (c) trained without BSW loss and (d) trained with BSW loss.}
							\label{fig:loss}
							\vspace{-8pt}
						\end{figure}

						\subsubsection{Analysis of Background Selective Whitening Loss}
						To verify the effectiveness of the proposed BSW loss, we visualized the distribution of the query features, which were learned with and without the background selective whitening loss, as shown in Fig.~\ref{fig:tsne}. 
						Specifically, we project the embedded content features from the test images into 2D space using t-SNE~\cite{van2014accelerating}.
						The color indicates	the memory items, which means that points with the same color are mapped to the same item. 
						The BSW loss was effective in separating and clustering the feature semantically. 
						Therefore, it enhances the diversity and discriminative power of our memory items.
						With the BSW loss, our method can represent various types of rain streaks.
						
						We further observed the effect of the background selective whitening loss shown in Fig.~\ref{fig:loss}.
						When our model was trained without $\mathcal{L}_{w}$, it was difficult to discriminate rain and background, as shown in Fig.~\ref{fig:loss} (c), whereas our model trained with $\mathcal{L}_{w}$ shows that the background information was removed, thus the rain streaks are captured in the memory only in Fig.~\ref{fig:loss} (d).
						This demonstrates that $\mathcal{L}_{w}$ helps to estimate rain effectively, and yields improved de-raining performance.
						
						Table~\ref{tab:abla} shows the evaluation of our model trained with memory networks in terms of various loss functions.
						The model trained with a BSW loss achieves better results than the model trained without the BSW loss, demonstrating the BSW loss helps de-raining.


						\begin{table}
							\centering
							\caption{
								\textbf{Quantitative evaluation of the proposed method trained on \textbf{RealDataset} and \textbf{TimeLap}, respectively.}
								Note that ($\cdot$) denotes the number of the scene.}
							\vspace{-5pt}
							\renewcommand{\tabcolsep}{4mm}
							\scalebox{0.9}{
								\begin{tabular}[b]{cc|cc}
									\toprule
									\textbf{RealDataset}  & \textbf{TimeLap} & PSNR  & SSIM  \\  
									\midrule
									\cmark (170) &  &  41.43  &  0.987 \\
									&  \cmark (170)  &  41.49   &  0.987 \\
									&  \cmark (186)  &  41.56   & 0.989  \\
									\cmark (170)   &  \cmark (186)  &  42.12   &  0.990 \\
									\bottomrule
								\end{tabular}}
								\label{tab:data}
								\vspace{-8pt}
							\end{table}

							\begin{table*}
								\centering
								\caption{
									\textbf{Comparison of run time (s) and the number of parameters.}
									Note that SPANet~\cite{SPANet} uses the CuPy library and thus can only be run on a GPU.}
								\vspace{-5pt}
								\renewcommand{\tabcolsep}{3.3mm}
								\scalebox{0.78}{   
									\begin{tabular}[b]{c|ccccccccccccc}
										\toprule
										&  DSC~\cite{luo2015removing} & GMM~\cite{luo2015removing} & JCAS~\cite{gu2017joint} & DDN~\cite{fu2017removing} & DID~\cite{zhang2018density} & JORDER-E~\cite{yang2019joint} & PReNet~\cite{ren2019progressive} & SIRR~\cite{SIRR} & SPANet~\cite{SPANet} & RCDNet~\cite{wang2020model} & Ours \\\cline{1-12}
										\midrule 
										CPU & 198.32 & 681.81 & 587.46 & 3.21 & 77.24 & 207.03 & 95.66 &  3.51  &  -- &  34.55 & 2.86  \\
										GPU & -- & -- & -- & 0.34 & 0.77 & 1.74 & 0.69 & 0.47  & 0.43 & 0.57 & 0.82  \\
										Params. & -- & -- & -- & 58.2K & 0.37M & 4.17M & 0.17M & 0.18M  & 0.28M &  3.17M &  0.81M \\
										
										\bottomrule
									\end{tabular}
								}
								\label{tab:runtime}
								\vspace{-11pt}
							\end{table*}

							

							\subsubsection{Analysis of Using Time-lapse Data}
							We conducted an experiment to compare the performance of the datasets according to the provided time-lapse  benchmark~\cite{SPANet,cho2020single}.
							Because \textbf{RealDataset} provides up to 170 scenes, for a fair experimental setup, we also selected 170 scenes of \textbf{TimeLab} randomly. Both data utilized two input pairs sampled from 30 images as training data in 170 scenes (\textit{i.e.,} 170 $\times$ $_{30}C_2$) for training.
							Table~\ref{tab:data} shows that the results trained with each dataset showed similar performance improvements because both datasets were constructed in real-world environments.
							Furthermore, we used the \textbf{TimeLab} to compare an experiment by varying the variety of scenes such that 170 scenes and 186 scenes (the total amount of data provided by \textbf{TimeLab}).
							Moreover, we used both datasets for training. From the results, we expect that the proposed method can achieve performance improvements when the time-lapse data contain various scenes.

							\begin{figure}
								\centering
								\renewcommand{\thesubfigure}{}
								\subfigure[(a)]{\includegraphics[width=0.115\textwidth]{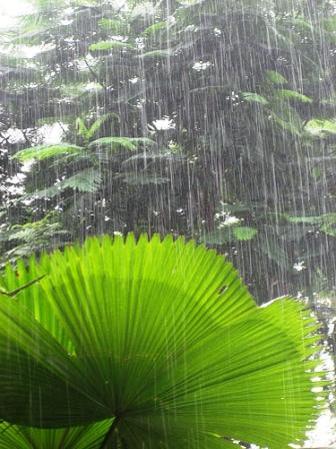}}
								\subfigure[(b)]{\includegraphics[width=0.115\textwidth,height=0.115\textheight]{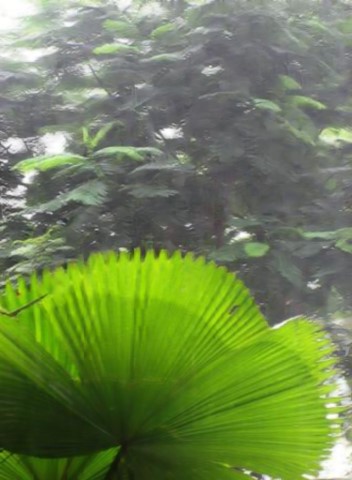}}
								\subfigure[(c)]{\includegraphics[width=0.115\textwidth,height=0.115\textheight]{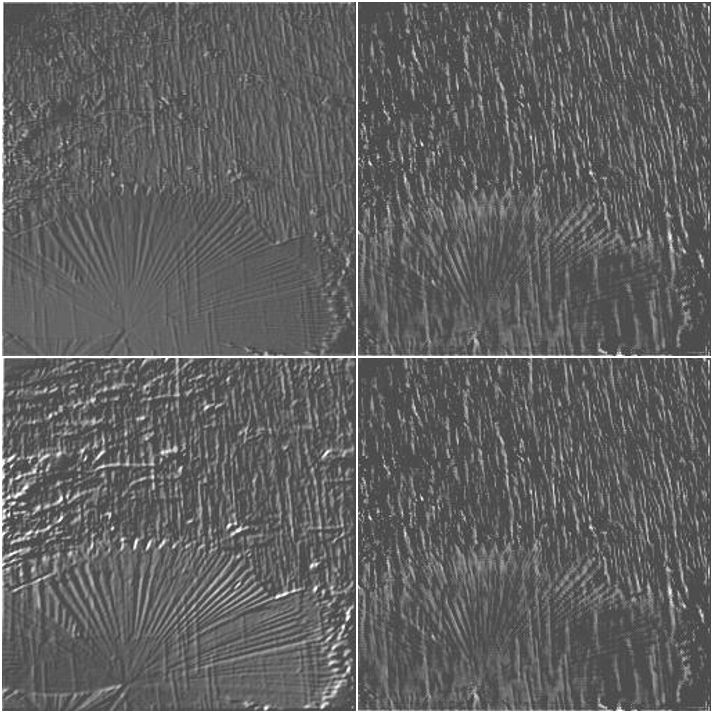}}
								\subfigure[(d)]{\includegraphics[width=0.115\textwidth,height=0.115\textheight]{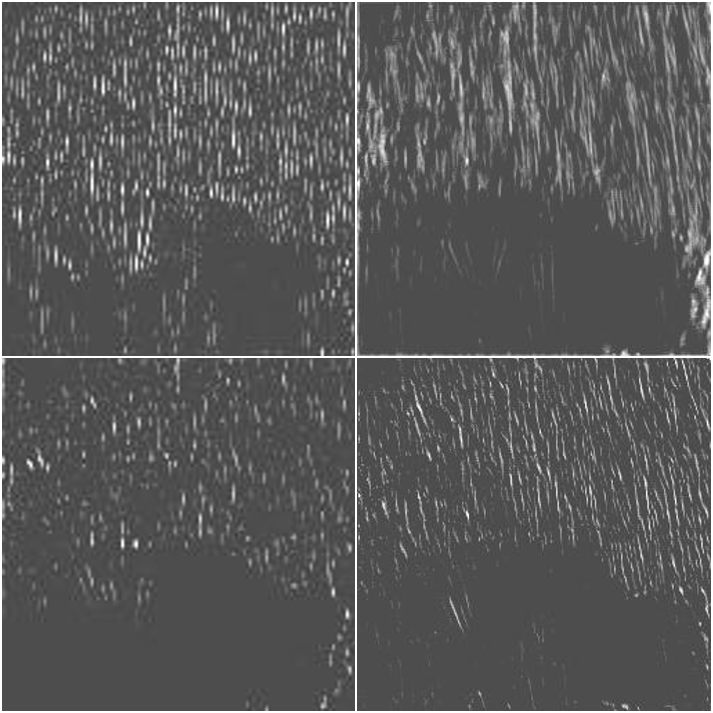}}\\
								\vspace{-3pt}
								\caption{
									\textbf{
										Visualization of de-raining and decoder features.} (a) Input rain image, (b) de-raining result,  four intermediate feature maps (c) in the encoder of the first convolution layers , and (d) from the decoder of the last convolution layers.}
								\label{fig:feat}
								\vspace{-13pt}
							\end{figure}

							\subsection{Visualization of Learned Features}
							In this section, we present the results of a visualization of learned features within our proposed networks.
							Fig.~\ref{fig:feat} shows a real rain image, a de-rained result, and visualizations of the learned feature maps of the first and last convolution layers.
							Fig.~\ref{fig:feat} (c) shows four intermediate feature maps in the encoder of the first convolution layers.
							It may clearly be observed that the first convolution seems to calculate the image gradients because the texture details of the leaf that were uncorrelated to the rain streaks are preserved.
							This indicates that the shallow layers mainly capture the image details extracted from the input rain image.
							In contrast, the feature maps of the last convolution layer presented in Fig.~\ref{fig:feat} (d) showed a high correlation with rain streaks, and contained various rain streaks.
							To summarize, this visualization demonstrates that the de-raining networks effectively estimate the rain streaks and whiten the background, generating plausible de-raining results.

							\subsection{Analysis of Run Time and Parameters}
							We provide data from the running time comparisons of our method with different existing methods in Table~\ref{tab:runtime}.
							The running times were averaged over 100 images with a size of 1000 $\times$ 1000 for evaluation. 
							The hand--crafted methods~\cite{luo2015removing,luo2015removing,gu2017joint} were run on the CPU according to the provided code, whereas other CNN--based methods were tested on both CPU and GPU.
							Our method shows GPU runtime similar to that of other other CNNs-based methods, and is much faster than most deep models on the CPU. Although the proposed method leverages the memory network, our model achieved improved de-raining results with comparable computational time.
							

							\section{Conclusion}
							
							We have presented a novel architecture for single image de-raining based on a memory network that can handle the various rain streak feature representations, thereby fully exploiting long-term rain streak information in the time-lapse data.
							We have also proposed a new background selective whitening (BSW) loss function designed to train a memory network to capture only rain streaks effectively by removing the consistent background information from the time-lapse data.
							The ablation studies clearly demonstrate the effectiveness of each component and the loss function in our framework.
							Extensive experiments also show that the proposed method achieved an improved generalization ability on both real-world and synthetic data.
							In future work, we will explore video de-raining by leveraging a memory network.

						\end{document}